%% file: main.tex
\documentclass{article} % For LaTeX2e
\usepackage{graphicx}
\usepackage{iclr2025_conference,times}

\input{math_commands.tex}

\usepackage{xcolor}
\definecolor{darkblue}{rgb}{0, 0, 0.5}

\usepackage[
    colorlinks=true,
    linkcolor=darkblue,
    citecolor=darkblue,
    urlcolor=blue
]{hyperref}

\usepackage{url}
\usepackage[capitalize]{cleveref}
\crefname{section}{Sec.}{Secs.}
\Crefname{section}{Section}{Sections}
\Crefname{table}{Table}{Tables}
\crefname{table}{Tab.}{Tabs.}

\usepackage{multirow}
\usepackage{rotating}
\usepackage[table]{xcolor}
\usepackage{tabularx}
\usepackage{arydshln}
\usepackage{siunitx, booktabs}
\usepackage{wrapfig} 

\usepackage{dblfloatfix}
\usepackage{fontawesome}
\usepackage{array}

\usepackage{booktabs}
\usepackage{multirow}
\usepackage[T1]{fontenc}
\usepackage[table,RGB]{xcolor}
\usepackage{hhline}
\usepackage{mysymbols}
\usepackage{authblk}   % ← this re‑defines \maketitle

\usepackage{makecell}
\usepackage{multirow}

\usepackage{enumitem}
\usepackage[normalem]{ulem}

\newlist{compactnum}{enumerate}{1}
\setlist[compactnum]{
    label=\arabic*.,
    leftmargin=2em,
    itemsep=0pt,
    parsep=0pt,
    topsep=0pt
}

\newcommand{\atokenbold}{\mbox{\textbf{AToken}}}
\newcommand{\atoken}{\mbox{\sc{AToken}}}
\newcommand{\atokensod}{\mbox{\sc{AToken}}-So/\textsc{d}}
\newcommand{\atokensoc}{\mbox{\sc{AToken}}-So/\textsc{c}}

\newcommand{\tablefont}{\small}

\usepackage{graphicx}
\usepackage{caption}
\usepackage{subcaption}

\usepackage{amssymb}% http://ctan.org/pkg/amssymb
\usepackage{pifont}% http://ctan.org/pkg/pifont
\definecolor{darkgreen}{RGB}{0, 150, 0} 
\newcommand{\cmark}{\textcolor{darkgreen}{\ding{51}}}
\newcommand{\xmark}{\textcolor{red}{\ding{55}}}%

\iclrfinalcopy % Uncomment for camera-ready version, but NOT for submission.

\title{AToken: A Unified Tokenizer for Vision}

\author{
\textbf{Jiasen Lu}$^{*}$
\quad  \textbf{Liangchen Song}$^{*}$ \quad \textbf{Mingze Xu} \quad \textbf{Byeongjoo Ahn} \\ \textbf{Yanjun Wang} \quad \textbf{Chen Chen} \quad  \textbf{Afshin Dehghan} \quad \textbf{Yinfei Yang} \\
\vspace{0.1in}
Apple
}

\crefname{table}{Table}{Tables}
\Crefname{table}{Table}{Tables}

\crefname{figure}{Figure}{Figures}
\Crefname{figure}{Figure}{Figures}

\crefname{section}{Section}{Sections}
\Crefname{section}{Section}{Sections}

\begin{document}

\renewcommand{\thefootnote}{*}
\footnotetext{Leading authors, equal contribution. Description of each author's contribution is available in Appendix~\ref{appendix:contribution}.
Corresponding to \texttt{Jiasen Lu}.
}

\maketitle

\input{sec/0_abstract}    
\input{sec/1_intro}

\input{sec/2_background}
\input{sec/3_method}
\input{sec/4_result}
\input{sec/5_application}
\input{sec/6_relate}
\input{sec/7_conclusion}

\appendix
\input{sec/supp_0_contribution}
% \input{sec/supp_1_expdetails}
% \input{sec/supp_2_figures}

% \newpage

{
\small
\bibliographystyle{main}
\bibliography{main}
}

\end{document}

%% file: math_commands.tex
\usepackage{amsmath,amsfonts,bm}

\def\eqref#1{equation~\ref{#1}}
\def\1{\bm{1}}

\DeclareMathAlphabet{\mathsfit}{\encodingdefault}{\sfdefault}{m}{sl}
\SetMathAlphabet{\mathsfit}{bold}{\encodingdefault}{\sfdefault}{bx}{n}

\usepackage{amsthm}

%% file: sec/0_abstract.tex
\begin{abstract}
We present \atoken, the first unified visual tokenizer that achieves both high-fidelity reconstruction and semantic understanding across images, videos, and 3D assets.
Unlike existing tokenizers that specialize in either reconstruction or understanding for single modalities, \atoken~encodes these diverse visual inputs into a shared 4D latent space, unifying both tasks and modalities in a single framework.
Specifically, we introduce a pure transformer architecture with 4D rotary position embeddings to process visual inputs of arbitrary resolutions and temporal durations.
To ensure stable training, we introduce an adversarial-free training objective that combines perceptual and Gram matrix losses, achieving state-of-the-art reconstruction quality.
By employing a progressive training curriculum, \atoken~gradually expands from single images, videos, and 3D, and supports both continuous and discrete latent tokens.
\atoken~achieves 0.21 rFID with 82.2\% ImageNet accuracy for images, 3.01 rFVD with 40.2\% MSRVTT retrieval for videos, and 28.28 PSNR with 90.9\% classification accuracy for 3D. 
In downstream applications, \atoken~enables both visual generation tasks (\eg, image generation with continuous and discrete tokens, text-to-video generation, image-to-3D synthesis) and understanding tasks (\eg, multimodal LLMs), achieving competitive performance across all benchmarks. 
These results shed light on the next-generation multimodal AI systems built upon unified visual tokenization.
\end{abstract}

%% file: sec/1_intro.tex
\section{Introduction}
\label{sec:introduction}

% paragraph 1: 
Large Language Models (LLMs) \citep{Chowdhery2022PaLMSL, achiam2023gpt, Touvron2023LLaMAOA, team2023gemini, guo2025deepseek} have achieved unprecedented generalization, with single models handling coding, reasoning, translation, and numerous other tasks that previously required specialized systems. 
This versatility largely stems from transformer architectures and simple tokenizers, such as BPE \citep{sennrich2015neural}, which convert all text types -- code, documents, tables, and multiple languages -- into a unified token space.
This shared representation enables efficient scaling and seamless knowledge transfer across language tasks.

% paragraph 2: 
In contrast, visual representations remain fragmented due to inherent complexities. 
Unlike text's discrete symbolic nature, visual tasks demand distinct levels of abstraction: generation requires tokenizers that preserve low-level visual details for reconstruction, while understanding requires encoders that extract high-level semantic features through text alignment.
Moreover, visual data exists in disparate formats: 2D grids for images, temporal sequences for videos, and varied 3D representations (\eg, meshes, voxels, and Gaussian splats) \citep{mescheder2019occupancy, achlioptas2018learning, mildenhall2021nerf, kerbl20233d}.
Without a shared representation, vision systems remain fundamentally limited, unable to achieve the generalization and transfer learning that characterizes modern language models.

% paragraph 3: 
Despite recent progress, unified visual tokenizers face three fundamental challenges. 
First, existing approaches optimize for either reconstruction or understanding, but not both: visual encoders \citep{radford2021learning, Zhai2023SigmoidLF, bolya2025perception} achieve semantic alignment but lack pixel-level detail, while VAE-based tokenizers \citep{esser2020taming, rombach2022high, polyak2024movie, Yu2022MAGVITMG} preserve visual details but lack semantic understanding. 
Second, architectural choices create different limitations: convolutional tokenizers exhibit diminishing returns when scaling model parameters \citep{xiong2025gigatok}, while transformer tokenizers \citep{yu2021vector,wang2024omnitokenizer,hansen2025learnings} achieve better scaling but suffer from severe adversarial training instabilities.
Third, recent unification efforts remain limited to images \citep{deng2025bagel,wu2024vila,ma2025unitok}, while video and 3D modalities remain unexplored.

\begin{figure}[t]
    \centering
    \includegraphics[width=1.0\linewidth]{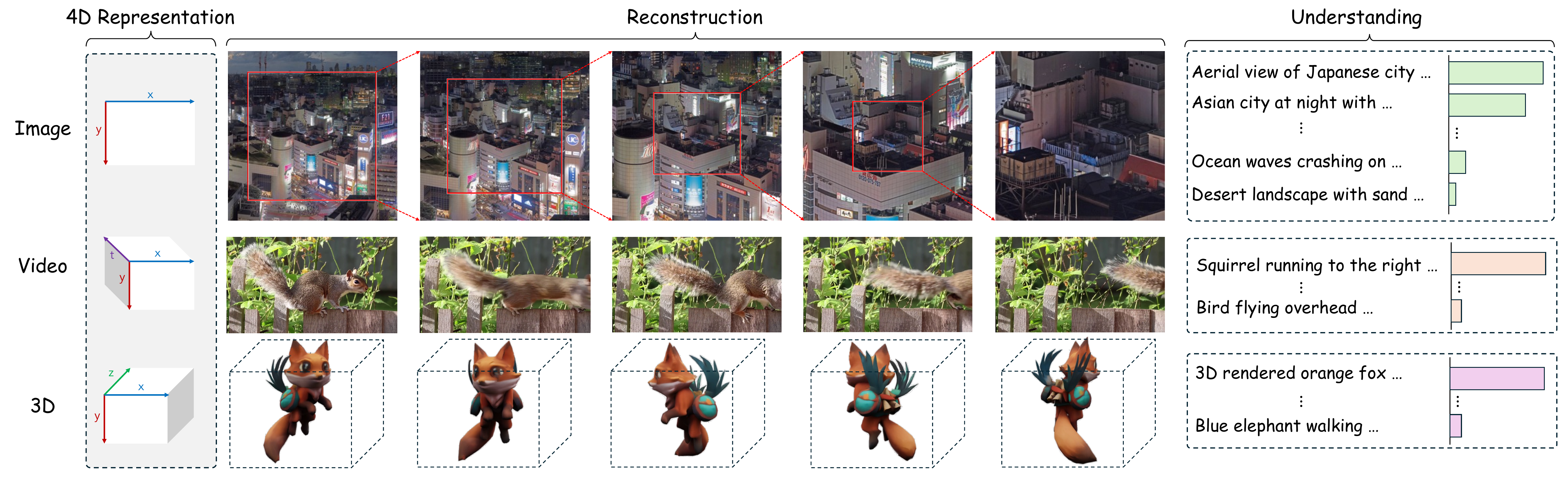}
    \caption{\small \textbf{Illustration of our method on different visual modalities.} Given images, videos, and 3D assets, \atoken~leverages a shared 4D latent space (left) to produce high-fidelity reconstructions (middle: zoomed regions with red boxes for images, temporal frames for videos, multiple viewpoints for 3D) while preserving strong semantic understanding (right: showing text-aligned representations for zero-shot text retrieval).}
    \label{fig:teaser_figure}
    \vspace{-3mm}
\end{figure}

% paragraph 4: In this paper 
In this paper, we present \atoken, a general-purpose visual tokenizer that achieves \textit{high-fidelity reconstruction} and \textit{rich semantic understanding} across \textit{images}, \textit{videos}, and \textit{3D}. 
Our model learns a unified representation that captures both fine-grained visual details and high-level semantics, accessible through progressive encoding: semantic embeddings for understanding, low-dimensional continuous latents for generation, and discrete tokens via quantization. 
This design enables the next generation of multimodal systems that seamlessly handle both understanding and generation across all visual modalities, as shown in \cref{fig:teaser_figure}.

% paragraph 5: Challenge 1
To address format discrepancies across visual modalities, we introduce a sparse 4D representation where each modality naturally occupies different subspaces: images as 2D slices, videos as temporal stacks, and 3D assets as surface voxels extracted from multi-view renderings \citep{xiang2024structured}. 
We implement this through a pure transformer architecture with space-time patch embeddings and 4D Rotary Position Embeddings (RoPE), enabling efficient scaling and joint modeling across all modalities while maintaining native resolution and temporal length processing.

% paragraph 6: Challenge 2
To overcome training instabilities that affect transformer-based visual tokenizers, we develop an adversarial-free loss combining perceptual and Gram matrix terms.
This approach achieves state-of-the-art reconstruction quality while maintaining stable, scalable training. 
We further introduce a progressive curriculum that builds capabilities incrementally: starting from a pretrained vision encoder, jointly optimizing reconstruction and understanding for images, extending to videos and 3D data, with optional quantization for discrete tokens. Surprisingly, this curriculum reveals that multimodal training can enhance rather than compromise single-modality performance -- our final model achieves better image reconstruction than earlier image-only stages while maintaining strong semantic understanding.

% paragraph 7: what atoken can do. 
\atoken~demonstrates significant advances in both scalability and performance. 
The model natively processes arbitrary resolutions and time duration, and accelerates inference through KV-caching mechanisms.
To validate its effectiveness, we conduct comprehensive evaluations across three dimensions: reconstruction quality, semantic understanding, and downstream applications.
These experiments confirm that \atoken~achieves competitive or state-of-the-art performance across all modalities while maintaining computational efficiency.

% contributions: 
The key contributions of \atoken~can be summarized as follows:
\begin{itemize}[leftmargin=*,itemsep=1pt,topsep=2pt]
\item \textbf{First unified visual tokenizer across modalities and tasks}: We present the first tokenizer that achieves high-fidelity reconstruction and semantic understanding for images, videos, and 3D assets, supporting both continuous and discrete representations within a single framework.

\item \textbf{Sparse 4D representation with pure transformer architecture}: We introduce a unified 4D latent space where different modalities naturally occupy respective subspaces, implemented through space-time patch embeddings and 4D RoPE that enable native resolution and temporal processing.

\item \textbf{Adversarial-free training for stable optimization}: We demonstrate that combining perceptual and Gram matrix losses achieves state-of-the-art reconstruction quality without adversarial training, overcoming instabilities that challenge transformer-based visual tokenizers.

\item \textbf{Progressive curriculum across modalities}: Our four-stage training strategy enables stable learning while maintaining strong performance, with image reconstruction quality preserved or improved when video and 3D capabilities are added alongside semantic understanding.

\item \textbf{Strong empirical validation across downstream applications}: \atoken~achieves competitive performance across all modalities and enables diverse applications from multimodal LLMs to image-to-3D generation, validating its effectiveness as a universal visual foundation.
\end{itemize}

%% file: sec/2_background.tex
\section{Background}
\label{sec:background}

Visual tokenization transforms raw visual data into compact representations suitable for both understanding and generation tasks. However, existing approaches remain fragmented across modalities and task objectives, unable to achieve the versatility seen in language models. \cref{tab:tasks} summarizes the landscape of visual tokenizers across three key dimensions: task specialization, modality fragmentation, and architectural trade-offs. A comprehensive review of related work is in \cref{sec:related_work}.

\paragraph{Task Specialization.}

Current visual tokenizers fall into two distinct categories based on their optimization objectives. 
Reconstruction methods like SD-VAE \citep{rombach2022high}, VQGAN \citep{esser2020taming}, GigaTok \citep{xiong2025gigatok}, and Cosmos \citep{Agarwal2025CosmosWF} excel at compressing visual data for generation tasks but cannot extract semantic features for understanding.
Conversely, understanding-centric visual encoders such as CLIP \citep{radford2021learning}, SigLIP2 \citep{tschannen2025siglip}, and VideoPrism \citep{zhao2024videoprism} produce rich semantic representations but cannot reconstruct the original visual content. Only recent works VILA-U \citep{wu2024vila} and UniTok \citep{ma2025unitok} attempt both tasks simultaneously, though they remain limited to images. This divide prevents building visual models that excel at both generation and understanding.

\paragraph{Modality Fragmentation.}
Beyond task specialization, visual tokenizers are limited to specific modalities. While most video tokenizers naturally handle images as single-frame videos (\eg, TAE \citep{polyak2024movie}, Hunyuan \citep{kong2024hunyuanvideo}, OmniTokenizer \citep{wang2024omnitokenizer}), they cannot process 3D data. Conversely, 3D tokenizers like Trellis-SLAT \citep{xiang2024structured} are restricted to 3D-only data, unable to leverage the massive image and video data for pretraining. Understanding tasks face similar constraints: image encoders process videos frame-by-frame without temporal compression, while dedicated video encoders \citep{zhao2024videoprism, Wang2022InternVideoGV} lack image-specific optimizations. No existing method provides comprehensive coverage across all three modalities for both reconstruction and understanding tasks.

\input{table/tasks}

\paragraph{Architectural Trade-offs.}
Key design trade-offs emerge across methods:
\textit{(1) Architecture:} Understanding encoders use transformers, while reconstruction tokenizers favor convolutional architectures (\eg, SD-VAE \citep{rombach2022high}). Recent works explore hybrid (\eg, GigaTok \citep{xiong2025gigatok}) and pure transformer approaches (\eg, ViTok \citep{hansen2025learnings}), though the latter suffer from adversarial training instabilities.
\textit{(2) Token representation:} Methods choose between discrete tokens for LLM compatibility (\eg, VQGAN \citep{esser2020taming}) or continuous tokens for reconstruction quality (\eg, TAE \citep{polyak2024movie}), with few supporting both.
\textit{(3) Resolution handling:} Convolutional architectures naturally handle arbitrary resolutions, while among transformer-based approaches, only SigLIP2 \citep{tschannen2025siglip} supports native resolution processing.
\textit{(4) Training objectives:} GAN-based training dominates reconstruction tokenizers for quality despite instabilities. Trellis-SLAT \citep{xiang2024structured} avoids adversarial training as 3D assets lack the fine detail of real images and videos.

These limitations motivate \atoken, which unifies reconstruction and understanding across images, videos, and 3D within a single transformer framework. As shown in \cref{tab:tasks}, \atoken~is the only method providing full coverage -- both tasks, all modalities, both token types -- while achieving training stability through adversarial-free optimization.

%% file: table/tasks.tex
\begin{table}[t]
\tablefont
\setlength\tabcolsep{5pt}
\renewcommand{\arraystretch}{1.15}
\newcolumntype{M}[1]{>{\centering\arraybackslash}p{#1}}
    \centering
    \caption{\small \textbf{Comparison between existing visual tokenizers and \atokenbold.} We categorize methods by task capabilities (reconstruction, understanding, or both) and evaluate their modality coverage, architectural choices, token representations, and key features. \atoken~is the only method providing support across all dimensions.}
    \label{tab:tasks}
      \resizebox{\columnwidth}{!}{
    \begin{tabular}{l cc M{1.cm}M{1.cm}M{1.cm} M{1.cm}M{1.cm}M{1.cm} ccccc}
        \toprule
        \multirow{3}{1.6cm}{Method} & \multirow{3}{1cm}{\centering Encoder Arch.} & \multirow{3}{1cm}{\centering Decoder Arch.} & \multicolumn{3}{c}{Reconstruction} & \multicolumn{3}{c}{ Understanding} & \multirow{3}{1cm}{\centering Discrete Token} & \multirow{3}{1cm}{\centering Cont. Token} & \multirow{3}{1.cm}{\centering GAN Free} & \multirow{3}{1.1cm}{\centering Temporal Comp.} & \multirow{3}{1.cm}{\centering Native Res.} \\
        \cmidrule(rl){4-6}
        \cmidrule(rl){7-9}
         & & & {Image} & { Video} & { 3D} & { Image} & {Video} & {3D} & & & & & \\
        \midrule
        \multicolumn{13}{l}{\textit{Reconstruction Only}} \\
        \quad SD-VAE  & Conv & Conv & \cmark & \xmark &\xmark &\xmark &\xmark &\xmark & \xmark & \cmark & \xmark & \xmark & - \\
        \quad  VQGAN  & Conv & Conv & \cmark & \xmark &\xmark &\xmark &\xmark &\xmark & \cmark & \xmark & \xmark & \xmark & - \\
        \quad  GigaTok  & Hybrid & Hybrid & \cmark & \xmark &\xmark &\xmark &\xmark &\xmark & \cmark & \xmark & \xmark & \xmark & \xmark \\
        \quad OmniTokenizer & Trans & Trans & \cmark & \cmark & \xmark & \xmark & \xmark & \xmark &  \cmark & \cmark & \xmark & \cmark & \xmark \\
        \quad MAGVIT-v2 & Conv & Conv & \cmark & \cmark & \xmark & \xmark & \xmark & \xmark & \cmark & \xmark & \xmark & \cmark & -\\
        \quad Cosmos & Conv & Conv & \cmark & \cmark &\xmark &\xmark &\xmark &\xmark & \cmark & \cmark & \xmark & \cmark & - \\ 
        \quad ViTok & Trans & Trans & \cmark & \cmark & \xmark & \xmark & \xmark & \xmark &  \xmark & \cmark & \xmark & \cmark & \xmark \\
        \quad TAE  & Conv & Conv & \cmark & \cmark &\xmark &\xmark &\xmark &\xmark & \xmark & \cmark & \xmark & \cmark & - \\
        \quad Hunyuan  & Conv & Conv & \cmark & \cmark &\xmark &\xmark &\xmark &\xmark & \xmark & \cmark & \xmark & \cmark & - \\
        \quad Wan  & Conv & Conv & \cmark & \cmark &\xmark &\xmark &\xmark &\xmark & \xmark & \cmark & \xmark & \cmark & - \\ 
        \quad Trellis-SLAT & Trans & Trans & \xmark & \xmark  & \cmark & \xmark & \xmark & \xmark & \xmark & \cmark & \cmark & \xmark & - \\
        \midrule
        \multicolumn{13}{l}{\textit{Understanding Only}} \\
        \quad SigLIP2 & Trans & - & \xmark & \xmark & \xmark & \cmark & \cmark & \xmark & - & - & - & \xmark & \cmark \\
        \quad PE  & Trans & - & \xmark & \xmark & \xmark & \cmark & \cmark & \xmark & - & - & - & \xmark & \xmark \\
        \quad VideoPrism & Trans & - & \xmark & \xmark & \xmark & \cmark & \cmark & \xmark & - & - & - & \xmark & \xmark \\
        \quad InternVideo & Trans & - & \xmark & \xmark & \xmark & \cmark & \cmark & \xmark & - & - & - & \cmark & \xmark \\
        \midrule
        \multicolumn{13}{l}{\textit{Reconstruction \& Understanding}} \\
        \quad VILA-U & Trans & Conv & \cmark & \xmark & \xmark & \cmark & \cmark & \xmark & \cmark & \xmark & \xmark & \xmark & \xmark \\
        \quad UniTok & Trans & Hybrid & \cmark & \xmark & \xmark & \cmark & \cmark & \xmark & \cmark & \xmark & \xmark & \xmark & \xmark \\
        \midrule
        \quad \atoken & Trans & Trans & \cmark & \cmark & \cmark & \cmark & \cmark & \cmark & \cmark & \cmark & \cmark & \cmark & \cmark \\
        \bottomrule
    \end{tabular}}
    \vspace{-3mm}
\end{table}

%% file: sec/3_method.tex
\section{Model}

This section presents \atoken's architecture and training methodology. We first present our unified 4D representation that bridges all visual modalities (\cref{subsec:latent}) and the pure transformer architecture that processes these representations (\cref{subsec:architecture}). We then describe our adversarial-free training objectives for stable optimization (\cref{subsec:loss}) and our progressive curriculum that enables effective multimodal learning (\cref{subsec:training}),  followed by implementation details (\cref{subsec:implement_details}).

\subsection{Unified Latent Representation}
\label{subsec:latent}

\begin{figure}
    \centering
    \includegraphics[width=1.0\linewidth]{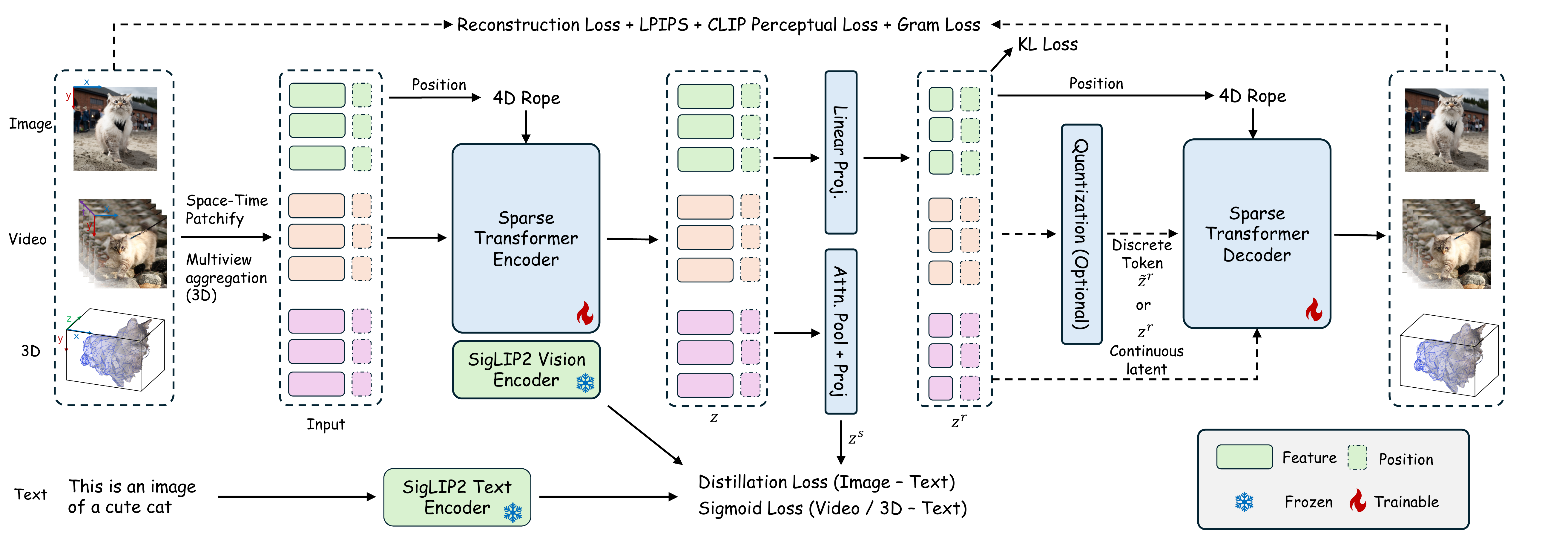}
    \caption{\small \textbf{Overview of our method.} All modalities undergo unified space-time patchification and encoding into sparse 4D latents, which support both reconstruction through modality-specific decoders and understanding through attention pooling and text alignment. The architecture jointly optimizes reconstruction and understanding losses, maintaining sparse structured representations throughout for efficient multimodal processing.}
    \label{fig:model}
    \vspace{-3mm}
\end{figure}

\paragraph{Unified Modalities -- Image, Video and 3D.}
Our central insight is that all visual modalities can be represented within a shared 4D space. As illustrated in \cref{fig:model}, we process each modality through space-time patchification to produce sets of feature-coordinate pairs:
\begin{equation}
    \bm{z} = \{(\bm{z}_i, \bm{p}_i)\}_{i=1}^L, \quad \bm{z}_i \in \mathbb{R}^C, \quad \bm{p}_i \in \{0,1,\ldots, N-1\}^4
\end{equation}
where $\bm{z}_i$ represents the latent feature at position $\bm{p}_i = [\mathrm{t}, \mathrm{x}, \mathrm{y}, \mathrm{z}]$ in 4D space (temporal and spatial coordinates), with $N$ defining the resolution along each axis and $L$ the number of active locations.

This sparse representation unifies all modalities by activating only their relevant dimensions: images occupy the $(\mathrm{x},\mathrm{y})$ plane at $\mathrm{t}=\mathrm{z}=0$, videos extend along the temporal axis with $\mathrm{z}=0$, and 3D assets as surface voxels in $(\mathrm{x},\mathrm{y},\mathrm{z})$ space with $\mathrm{t}=0$. For 3D assets, we adapt Trellis-SLAT \citep{xiang2024structured} by rendering multi-view images from spherically sampled cameras, applying our unified patchification, then aggregating features into voxel space (detailed in \cref{subsec:architecture}). This approach enables a single encoder $\mathcal{E}$ to process all modalities without architectural modifications.

Note that the $(\mathrm{x},\mathrm{y},\mathrm{z})$ coordinates serve different purposes across modalities: in 3D, they represent actual entity occupancy physical locations, while in images and videos, they function as grid indices. We can conceptualize this as placing a monitor within 4D space and encoding its displayed content for image and video data. This dual interpretation of coordinates does not compromise generalization, thanks to the use of 4D RoPE, which we describe in detail in following sections.

\paragraph{Unified Tasks -- Reconstruction and Understanding.} 
From the unified structured latents $\bm{z} = \{(\bm{z}_i, \bm{p}_i)\}$, we extract representations for both reconstruction and understanding through complementary projections.
For reconstruction, we project each latent to a lower-dimensional space $\bm{z}^r = \bm{W}_r(\bm{z})$ with KL regularization \citep{rombach2022high}, optionally applying FSQ \citep{Mentzer2023FiniteSQ} for discrete codes $\tilde{\bm{z}}^r = \text{FSQ}(\bm{z}^r)$. The decoder $\mathcal{D}_\theta$ then reconstructs the input from these latents.
For understanding, we aggregate latents via attention pooling \citep{radford2021learning, tschannen2025siglip} into a global representation $\bar{\bm{z}}$, which is projected to $\bm{z}^s = \bm{W}_s(\bar{\bm{z}})$ for alignment with text embeddings.
This dual projection design allows joint optimization without architectural duplication -- the same encoded features $\bm{z}$ support both pixel-level reconstruction through individual latents and semantic understanding through their aggregation.

\subsection{Transformer based Architecture}
\label{subsec:architecture}

\paragraph{Unified Space-Time Patch Embedding.} 
% How we patchify image and video
We employ a unified patchification scheme that enables all modalities to share the same encoder. Given an input $\bm{x} \in \mathbb{R}^{T \times H \times W \times 3}$, we partition it into non-overlapping space-time patches of size $t \times p \times p$. For images ($T=1$), we apply temporal zero-padding to create $t$-frame patches, ensuring consistent dimensions across modalities. Videos are directly partitioned along both spatial and temporal dimensions.

\begin{figure}
    \centering
    \includegraphics[width=1.0\linewidth]{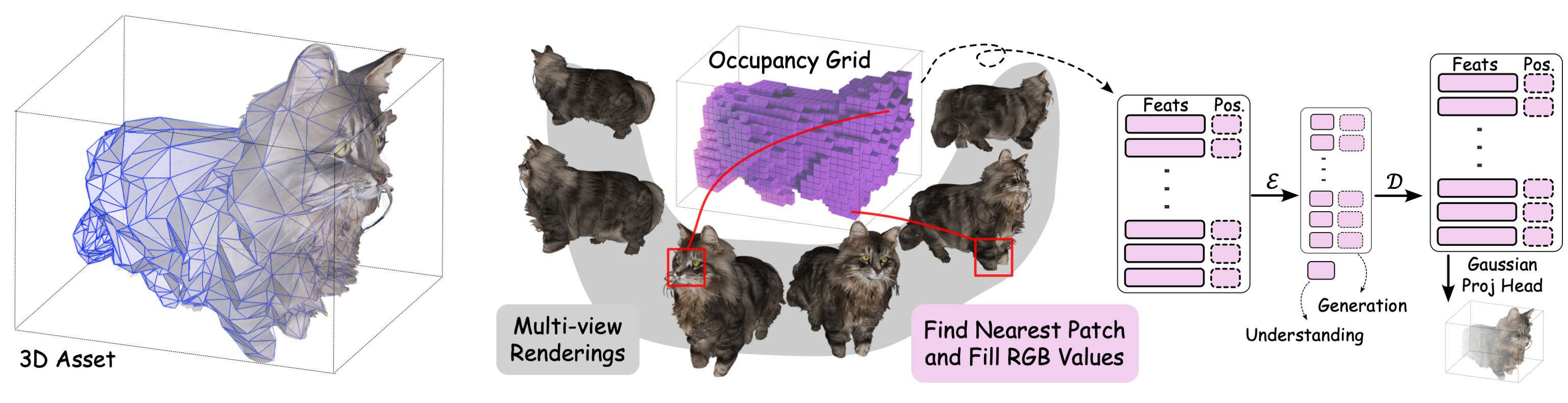}
    \caption{\small
    \textbf{3D tokenization pipeline.}
    We extend Trellis-SLAT \citep{xiang2024structured} for multimodal unification through two modifications: directly tokenizing raw RGB patches from multiview renderings (as opposed to using DINOv2 features), and aggregating each voxel's features from its nearest viewpoint (as opposed to averaging across all views). Combined with Gaussian decoding, this approach integrates 3D assets into our unified token space alongside images and videos.
    }
    \label{fig:mvproj}
    \vspace{-3mm}
\end{figure}

For 3D assets, we adapt Trellis-SLAT \citep{xiang2024structured} to our unified pipeline. As shown in \cref{fig:mvproj}, we render multi-view images from spherically sampled cameras and apply our standard space-time patchification. Each voxel in a $64^3$ grid is back-projected to gather and average patch features from relevant views. Unlike \cite{xiang2024structured}, which uses DINOv2 features, we achieve comparable quality using our unified patch representation. 

All patch features -- whether from images, videos, or aggregated 3D views -- are then flattened and passed through a shared linear layer to produce the initial embeddings for the transformer encoder.

\paragraph{Sparse Transformer Encoder and Decoder.}
We employ a unified transformer architecture for both encoder and decoder, as illustrated in \cref{fig:model}. Both components process sparse structured representations -- sets of feature-position pairs rather than dense grids -- enabling efficient handling of all modalities with native support for arbitrary resolutions and temporal lengths.

Our encoder $\mathcal{E}$ extends the pretrained SigLIP2 vision tower \citep{tschannen2025siglip} from 2D images to 4D representations through two modifications. First, we generalize patch embedding to space-time blocks of size $t \times p \times p$, with zero-initialized temporal weights preserving the original image features. Second, we augment SigLIP2's learnable 2D position embeddings with 4D RoPE \citep{lu2024unified} applied in every attention layer, providing relative position awareness across $(\mathrm{t},\mathrm{x},\mathrm{y},\mathrm{z})$ dimensions. This design maintains SigLIP2's semantic priors and resolution flexibility while enabling unified processing across modalities.

% % Decoder: 
The decoder $\mathcal{D}$ shares the encoder's transformer architecture but is trained from scratch for reconstruction. It maps structured latents back to visual outputs through task-specific heads. For images and videos, we decode directly to pixel space:
\begin{equation}
    \mathcal{D}_\mathrm{P}: \{(\bm{z}_i,\bm{p}_i)\}_{i=1}^{L} \rightarrow \bm{x} \in \mathbb{R}^{T \times H \times W \times 3}
\end{equation}
treating images as single-frame videos ($T=1$) and discarding temporal padding following \citep{polyak2024movie}. For 3D assets, we first decode to pixel-space features, then apply an additional layer to generate Gaussian splatting parameters for efficient rendering:
\begin{equation}
    \mathcal{D}_\mathrm{GS}: \{(\bm{z}_i,\bm{p}_i)\}_{i=1}^{L} \rightarrow \{\{(\bm{o}_i^k,\bm{c}_i^k,\bm{s}_i^k,\alpha_i^k,\bm{r}_i^k)\}_{k=1}^{K}\}_{i=1}^{L}
\end{equation}
where each location generates $K$ Gaussians with parameters: position offset $\bm{o}$, color $\bm{c}$, scale $\bm{s}$, opacity $\alpha$, and rotation $\bm{r}$. Following \cite{xiang2024structured}, we constrain Gaussian positions to remain near their source voxels using $\bm{x}_i^k=\bm{p}_i + \texttt{tanh}(\bm{o}_i^k)$, ensuring local feature coherence.

\subsection{Training Objectives}
\label{subsec:loss}

We jointly optimize for reconstruction fidelity and semantic understanding through an adversarial-free training objective:

\begin{equation}
\mathcal{L} = \lambda_{\text{rec}} \mathcal{L}_{\text{rec}} + \lambda_{\text{sem}} \mathcal{L}_{\text{sem}} + \lambda_{\text{KL}} \mathcal{L}_{\text{KL}},
\end{equation}
where $\mathcal{L}_{\text{KL}}$ is the KL regularization term applied to the projected reconstruction latents $\bm{z}^r$, with $\lambda_{\text{rec}}$, $\lambda_{\text{sem}}$ and $\lambda_{\text{KL}}$ balancing components. Notably, we achieve state-of-the-art reconstruction quality without adversarial training, which has been observed to be unstable when scaling \citep{wu2025qwenimagetechnicalreport} and incompatible with our sparse 3D representations.

\begin{figure*}[t!]
    \centering
    \begin{subfigure}[b]{0.3\textwidth}
        \centering
        \includegraphics[width=\linewidth]{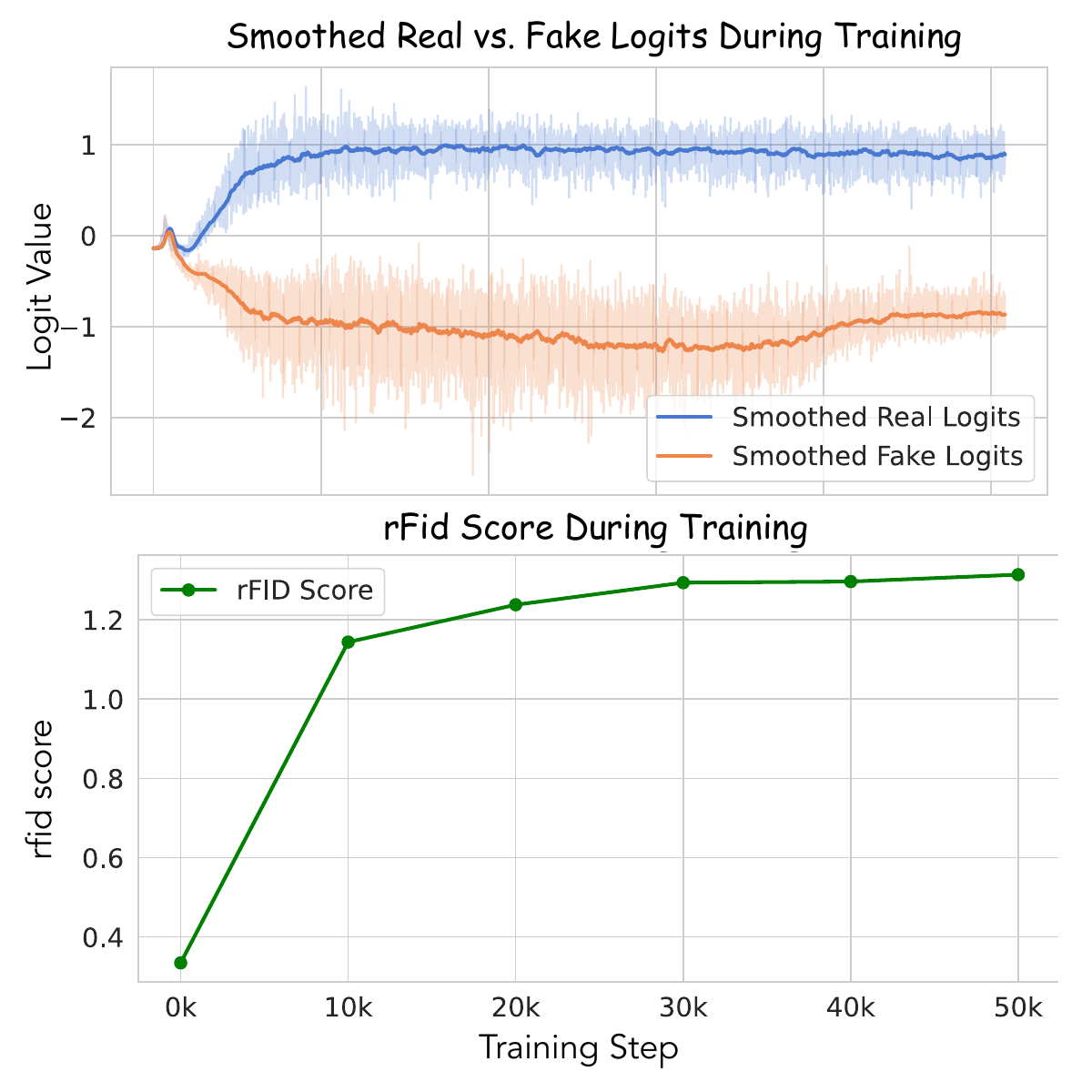}
        \caption{\footnotesize GAN training instability}
        \label{fig:gan_instability}
    \end{subfigure}
    \hfill
    \begin{subfigure}[b]{0.3\textwidth} % Narrower width for the bar plot
        \centering
        \includegraphics[width=\linewidth]{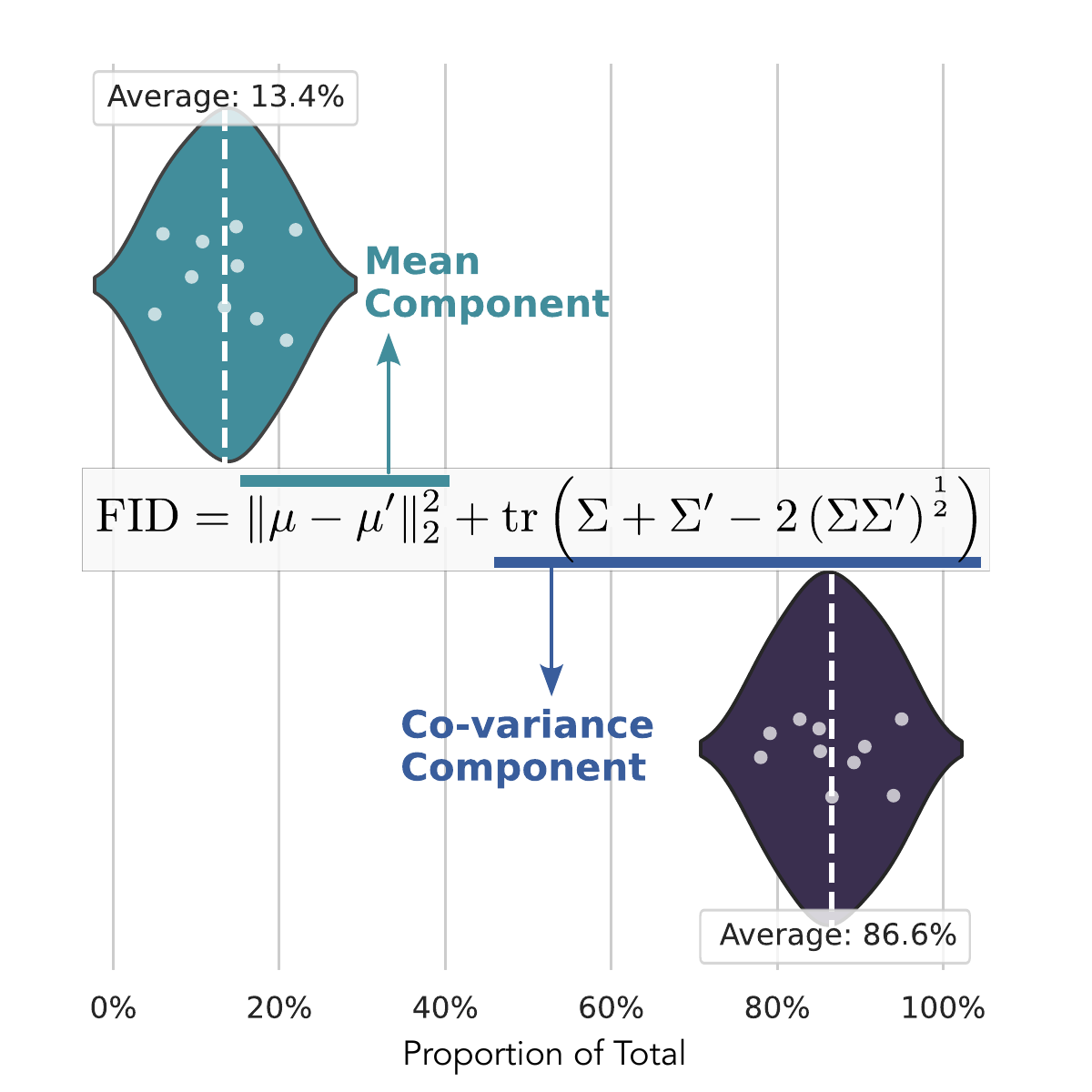}
        \caption{\footnotesize Decomposition of rFID.}
        \label{fig:rfid_decomposition}
    \end{subfigure}
    \hfill
    \begin{subfigure}[b]{0.3\textwidth}
        \centering
        \includegraphics[width=\linewidth]{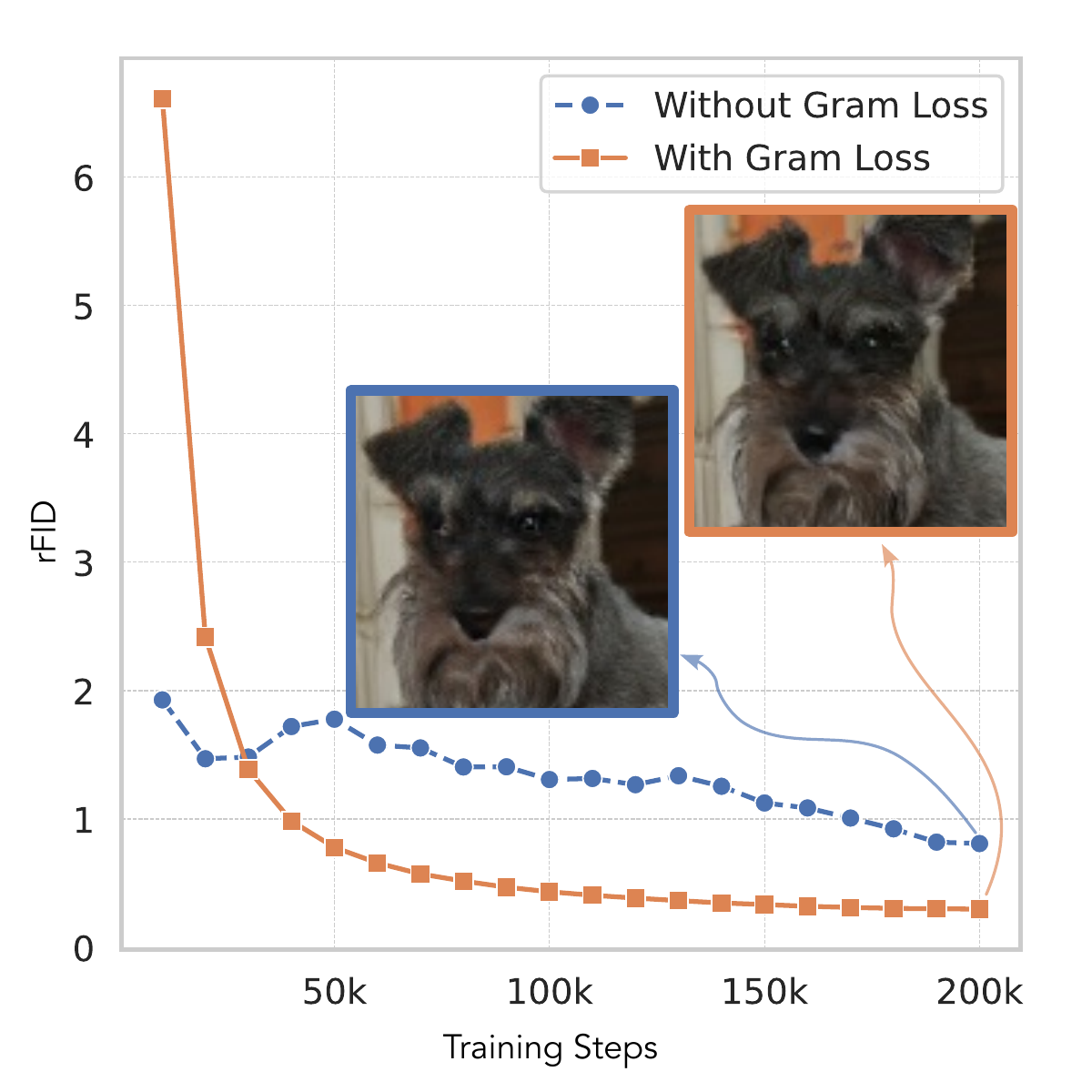}
        \caption{\footnotesize Gram loss efficiency}
        \label{fig:gram_efficacy}
    \end{subfigure}
\caption{\small \textbf{Adversarial-free training with Gram loss achieves stable, high-fidelity reconstruction.}
(a) GAN training fails in our setting: the discriminator overpowers the generator, causing diverging logits and degraded rFID.
(b) Decomposing rFID reveals $\approx86.6\%$ of error stems from covariance (texture/style) vs. $\approx13.4\%$ from mean components.
(c) Gram loss directly optimizes second-order statistics (\ie, feature covariance) without adversarial training, achieving superior and stable rFID throughout training.}
\label{fig:gram_loss_analysis}
\end{figure*}

\paragraph{Reconstruction Loss.}
While GANs \citep{goodfellow2014generative} are standard for visual tokenizers, we found them unsuitable for our transformer architecture. \cref{fig:gram_loss_analysis}(a) shows the discriminator rapidly dominates the generator, causing mode collapse and degraded reconstruction quality. To develop an alternative, we analyzed the reconstruction error by decomposing rFID into mean and covariance components (\cref{fig:gram_loss_analysis}(b)). The covariance component -- capturing second-order statistics like texture and style -- dominates at $\approx86.6\%$, while mean features contribute only 13.4\%. 

This insight motivated adopting Gram matrix loss \citep{Gatys2016ImageST}, which directly optimizes feature covariance without adversarial training:
\begin{equation}
\mathcal{L}_{\text{Gram}}(\bm{x}, \hat{\bm{x}}) = \sum_{l} \left\| G(\Phi_l(\bm{x})) - G(\Phi_l(\hat{\bm{x}})) \right\|_F^2,
\end{equation}
where $G(F) = FF^\top$ is the Gram matrix for feature map $F$ from layer $l$ of network $\Phi$. As shown in \cref{fig:gram_loss_analysis}(c), this achieves superior and stable reconstruction throughout training.

For images, we combine four complementary loss components:
\begin{equation}
\label{eq:rec_loss}
    \mathcal{L}_{\text{rec}}^{\text{I}} = \lambda_1 \mathcal{L}_1 + \lambda_{\text{LPIPS}} \mathcal{L}_{\text{LPIPS}} + \lambda_{\text{GRAM}} \mathcal{L}_{\text{GRAM}} + \lambda_{\text{CLIP}} \mathcal{L}_{\text{CLIP}},
\end{equation}
where $\mathcal{L}_1 = \|\bm{x} - \hat{\bm{x}}\|_1$ provides pixel supervision, $\mathcal{L}_{\text{LPIPS}}$ \citep{zhang2018perceptual} measures perceptual similarity, $\mathcal{L}_{\text{GRAM}}$ captures texture, and $\mathcal{L}_{\text{CLIP}}$ enforces semantic consistency. For video and 3D assets, we use $\mathcal{L}_{\text{rec}}^{\text{V/3D}} = \mathcal{L}_1$ for efficiency, relying on cross-modal transfer from images for details.

\paragraph{Semantic Loss.}
We align visual representations $\bm{z}^s$ with text embeddings through modality-specific objectives. For images, we distill knowledge from the frozen SigLIP2 vision encoder \citep{tschannen2025siglip} by minimizing the KL divergence between temperature-scaled vision-text similarity distributions:
\begin{equation}
    \mathcal{L}_{\text{sem}}^{\text{I}} = \text{KL}\left(\text{softmax}(\tau^{-1} \bm{s}^{\text{teacher}}) \,\|\, \text{softmax}(\tau^{-1} \bm{s}^{\text{student}})\right),
\end{equation}
where $\bm{s}^{\text{teacher}}$ and $\bm{s}^{\text{student}}$ are vision-text similarity scores from frozen SigLIP2 and our model respectively, both paired with the same frozen text encoder, and $\tau$ is the temperature parameter. 
For videos and 3D, we directly optimize alignment using the sigmoid loss from SigLIP \citep{Zhai2023SigmoidLF}, which proves more stable for the smaller batch sizes typical in these domains. This dual strategy preserves pretrained image semantics while enabling efficient learning for new modalities.

\begin{figure}[t]
    \centering
    \includegraphics[width=1.0\linewidth]{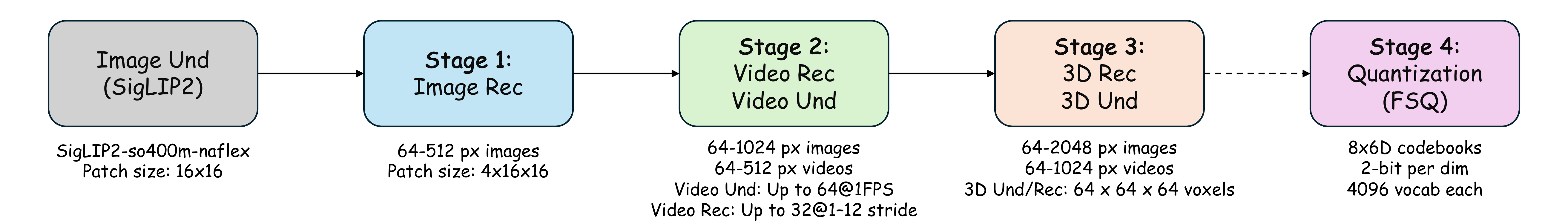}
    \caption{\small \textbf{Progressive training curriculum of \atokenbold.} Our model starts from SigLIP2 image understanding and progressively adds: (1) image reconstruction, (2) video capabilities with temporal modeling, (3) 3D understanding with expanded resolutions, and optionally (4) discrete tokenization via FSQ. Each box shows the new capabilities introduced at that stage, along with supported resolutions, patch sizes, and sampling strategies.}
\label{fig:stages}
\vspace{-3mm}
\end{figure}

\subsection{Training Strategy}
\label{subsec:training}

Our training employs a four-stage progressive curriculum (\cref{fig:stages}) that builds from image foundations to video dynamics to 3D geometry, with optional discrete quantization. Starting from the pretrained SigLIP2 encoder \citep{tschannen2025siglip}, we gradually introduce more complex objectives and modalities while maintaining semantic understanding across all stages.

We implement this curriculum through round-robin sampling of modalities and tasks, using gradient accumulation to balance image-text distillation with other objectives (reconstruction, video-text alignment, 3D-text alignment) across all stages. This ensures semantic alignment is preserved even as reconstruction capabilities expand. 
Our sparse transformer architecture facilitates this multi-modal training by separating features and positions, allowing each modality to be processed at its natural resolution without padding or packing. 

\paragraph{Stage 1: Image Foundation.}
Starting from pretrained SigLIP2, we establish core visual representations by adding image reconstruction capabilities. 
We process images using 4$\times$16$\times$16 space-time patches with temporal padding for consistency, employing 32 latent dimensions following \citep{Yao2025ReconstructionVG}. Training uses variable resolution sampling from 64 to 512 pixels, with L1 loss computed at native resolution while perceptual losses ($\mathcal{L}_{\text{LPIPS}}$, $\mathcal{L}_{\text{CLIP}}$, $\mathcal{L}_{\text{Gram}}$) use $224 \times 224$ interpolation to match their pretrained features.

\begin{figure}
    \centering
    \includegraphics[width=1.0\linewidth]{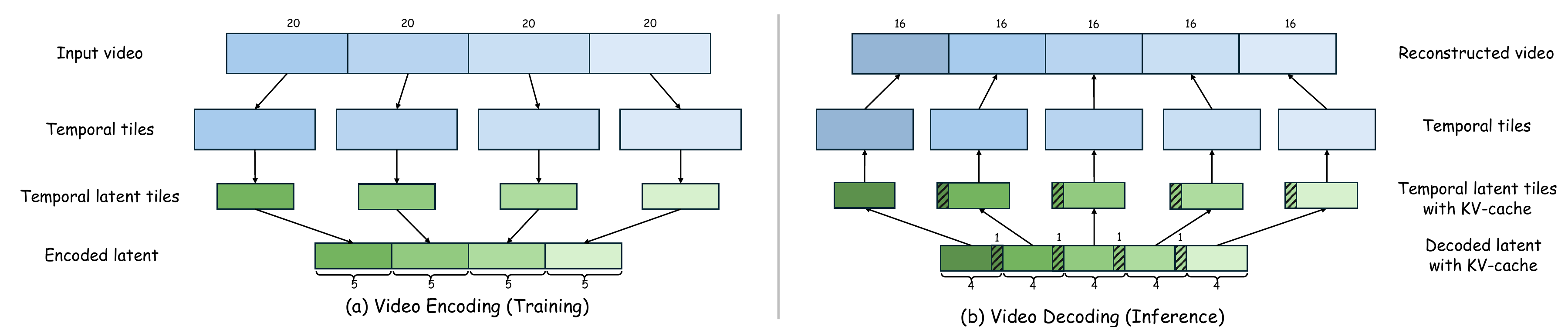}
    \caption{\small \textbf{Overview of the video encoding and decoding process.} During encoding, we use KV-caching across temporal tiles to eliminate redundant computation while maintaining temporal coherence, providing significant efficiency gains over overlapping tile methods.}
    \label{fig:video_encoding}
\end{figure}

\paragraph{Stage 2: Video Dynamics.}
We extend to temporal sequences, expanding latent dimensions from 32 to 48 to accommodate motion complexity \citep{Seawead2025Seaweed7BCT}. Resolution capabilities increase to 1024 for images and 512 for videos. We employ temporal tiling (16-32 frames $\rightarrow$ 4-8 latent frames) with adaptive sampling: stride 1-3 for temporal consistency or 4-12 for diversity in reconstruction, 1 FPS up to 64 frames for understanding. Our KV-caching mechanism (\cref{fig:video_encoding}) eliminates redundant computation across tiles while maintaining temporal coherence.

\paragraph{Stage 3: 3D Geometry.}
We incorporate 3D assets as active voxels in $64^3$ grids, using Gaussian splatting for reconstruction and attention pooling for understanding. Resolution further increases to 2048 for images and 1024 for videos. Joint optimization across all three modalities prevents catastrophic forgetting while leveraging cross-modal learning. The geometric semantics from 3D and the temporal dynamics from video enhance image reconstruction quality.

\paragraph{Stage 4: Discrete Tokenization.}
Optionally, we add FSQ quantization \citep{Mentzer2023FiniteSQ} for discrete generation tasks. The 48-dimensional latents are partitioned into 8 groups of 6 dimensions, each quantized to 4 levels, yielding 8 discrete tokens from 4096-entry codebooks. 
We finetune the entire encoder and decoder to adapt all modalities to discrete tokens, enabling compatibility with discrete generative models across all visual domains.

\input{table/training_stage}

\subsection{Implementation Details}
\label{subsec:implement_details}

Our encoder and decoder each contain 27 transformer blocks with hidden dimension $d=1152$ and 16 attention heads. The encoder is initialized from SigLIP-SO400M-patch16-naflex \citep{tschannen2025siglip}, while the decoder is trained from scratch. 

We optimize using AdamW with $\beta_1=0.9$, $\beta_2=0.95$, and weight decay $0.1$. The learning rate follows linear warmup for 2,000 steps to $\eta_{\max}=3\times10^{-4}$, then cosine annealing to $\eta_{\min}=3\times10^{-5}$. Given the pretrained encoder, we apply a reduced learning rate $\eta_{\text{encoder}} = 0.1 \times \eta_{\text{base}}$ and use exponential moving average with decay rate $\gamma=0.9999$. 

Training utilizes 256 H100 GPUs with adaptive global batch sizes optimized for each task's memory requirements. Image understanding maintains 8,192 samples throughout all stages, while reconstruction tasks scale with complexity: image reconstruction uses 1,024-4,096, video reconstruction uses 512-1024, and 3D reconstruction uses 256-512. The four-stage curriculum trains for 200k, 200k, 50k, and 100k iterations, respectively, with each stage initialized from the previous checkpoint, requiring a total of 138k GPU hours across all stages (approximately 22 days with 256 GPUs).

Throughout training, we maintain fixed loss coefficients: $\lambda_{\text{rec}}=0.2$, $\lambda_{\text{sem}}=1.0$, and $\lambda_{\text{KL}}=10^{-8}$. Within reconstruction (Eq.~\ref{eq:rec_loss}), we set $\lambda_1=1.0$, $\lambda_{\text{LPIPS}}=10.0$, $\lambda_{\text{GRAM}}=10^3$, $\lambda_{\text{CLIP}}=1.0$, and $\tau=2.0$. We normalize reconstruction losses over patches rather than summing \citep{esser2020taming}, providing stable gradients across resolutions. 

Training data follows our progressive curriculum: DFN \citep{Fang2023DataFN}, Open Images \citep{OpenImages}, and internal datasets for images; WebVid \citep{Bain21} and TextVR \citep{wu2025large} for video understanding with Panda70M \citep{chen2024panda} for reconstruction; Objaverse \citep{deitke2023objaverse} with Cap3D \citep{luo2024view} annotations for 3D. Datasets are sampled proportionally to their size, with task ratios detailed in Table~\ref{tab:training_stage}.

%% file: table/training_stage.tex
\begin{table}[t]
\tablefont
\setlength\tabcolsep{5pt}
\renewcommand{\arraystretch}{1.15}
\newcolumntype{M}[1]{>{\centering\arraybackslash}p{#1}}
    \centering
    \caption{\small \textbf{Training curriculum configuration.} Resolution limits for each modality and task sampling ratios across the four training stages. Superscripts denote reconstruction (r) and understanding (u) tasks.}

    \label{tab:training_stage}
      \resizebox{\columnwidth}{!}{
    \begin{tabular}{l ccc M{0.8cm}M{0.8cm}M{0.8cm}M{0.8cm}M{0.8cm} c }
        \toprule
        \multirow{3}{2.0cm}{Training Stage} & \multirow{3}{1.5cm}{\centering Image Res.} & \multirow{3}{1.5cm}{\centering Video Res.} & \multirow{3}{1.5cm}{\centering 3D Size} &  \multicolumn{5}{c}{Task Sampling Ratios}  &  \multirow{3}{1cm}{\centering \#Steps}\\
        % \multicolumn{3}{c}{Reconstruction} & \multicolumn{3}{c}{ Understanding} & \multirow{3}{1cm}{\centering Discrete Token} & \\
        % \cmidrule(rl){4-6}
        \cmidrule(rl){5-9}
        
        & & & & $\text{I}^r$ & $\text{V}^u$ &$\text{V}^r$ &$\text{3D}^u$ &$\text{3D}^r$ & \\
         % & & & {Image} & { Video} & { 3D} & { Image} & {Video} & {3D} & \\
        \midrule
        Stage 1: Image Foundation  &  $[64 \rightarrow~~512]$ & - & -  & 100\% & - & - & - & - & 200k\\
        Stage 2: Video Dynamics  &  $[64\rightarrow1024]$ & $[64 \rightarrow ~~512]$ & - & 22.2\% & 11.1\% & 66.6\% & - & - & 200k\\
        Stage 3: 3D Geometry  &  $[64 \rightarrow 2048]$ & $[64 \rightarrow 1024]$ & $[64,64,64]$  & 22.2\%  & 11.1\% & 44.4\% & 11.1\% & 11.1\% & 50k\\
        Stage 4: Discrete Tokenization  & $[64 \rightarrow 2048]$ & $[64 \rightarrow 1024]$ & $[64, 64, 64]$ & 22.2\%  & 11.1\% & 44.4\% & 11.1\% & 11.1\% & 100k\\
        \bottomrule
    \end{tabular}}
    \vspace{-3mm}
\end{table}

%% file: sec/4_result.tex
\section{Main Results}
We evaluate \atoken~as the first visual tokenizer to achieve both reconstruction and understanding across images, videos, and 3D assets. This section presents unified comparisons (\cref{subsec:unified_tokenizer}) followed by per-modality analysis (Sections~\ref{subsec:image}-\ref{subsec:3d}) and ablations (\cref{subsec:ablation}).

\subsection{Unified Tokenizer Comparisons}
\label{subsec:unified_tokenizer}
\input{table/main_results.tex}

\cref{tab:main_results} presents a comparison of visual tokenizers across modalities. We evaluate on standardized benchmarks: ImageNet \citep{deng2009imagenet} at 256$\times$256  (reconstruction: PSNR, rFID; understanding: zero-shot accuracy), TokenBench \citep{Agarwal2025CosmosWF} at 720p and MSR-VTT \citep{Xu2016MSRVTTAL} for video (reconstruction: PSNR, rFVD; understanding: text-to-video R@1), and Toys4k \citep{stojanov2021using} for 3D (reconstruction: PSNR, LPIPS; understanding: zero-shot accuracy).

The results reveal three distinct categories of approaches, each with fundamental limitations. Reconstruction-only tokenizers excel at generation but cannot extract semantic features: SD-VAE~\citep{rombach2022high}, FLUX.1~\citep{labs2025flux1kontextflowmatching}, VA-VAE~\citep{Yao2025ReconstructionVG}, and Qwen-Image~\citep{wu2025qwenimagetechnicalreport} for images; Hunyuan~\citep{kong2024hunyuanvideo} and WAN~\citep{wan2025} for video; Trellis-SLAT~\citep{xiang2024structured} for 3D. Understanding-only encoders provide rich semantics but cannot reconstruct visual content: SigLIP2~\citep{tschannen2025siglip}, VideoPrism~\citep{zhao2024videoprism}, and PE$_{\texttt{core}}$~\citep{bolya2025perception}. Recent unified attempts combine both capabilities but remain limited to images: SeTok~\citep{wu2024towards}, VILA-U~\citep{wu2024vila}, and UniTok~\citep{ma2025unitok}.

\atoken-So/\textsc{c} breaks these boundaries as the first tokenizer to unify all three capabilities. On images, we achieve 0.21 rFID with 82.2\% zero-shot ImageNet accuracy, substantially outperforming UniTok's 0.36 rFID and 78.6\% accuracy. More importantly, we extend this unified capability to video (3.01 rFVD, 40.2\% R@1) and 3D (28.28 PSNR, 90.9\% accuracy), comparable or even surpassing specialized methods like Wan2.2 and Trellis-SLAT on Video and 3D reconstruction. Our discrete variant (\atoken-So/\textsc{d}) maintains competitive performance, pioneering discrete tokenization across all modalities.

\input{table/image_rec_all.tex}

\subsection{Image Tokenization}
\label{subsec:image}
We evaluate \atoken's image capabilities against specialized tokenizers through reconstruction quality (\cref{tab:main_image_rec}) and semantic understanding (\cref{tab:main_image_zeroshot}) benchmarks.

\textbf{Reconstruction Performance.} 
\cref{tab:main_image_rec} presents our comprehensive evaluation, where we re-evaluated all baseline methods using a unified protocol with official implementations to ensure fair comparison. Under this standardized evaluation protocol, we observe that multimodal training enhances rather than compromises image reconstruction. \atoken-So/\textsc{c} achieves 0.209 rFID at 16$\times$16 compression, with progressive improvement across training stages: 0.258 (Stage 1)$\rightarrow$0.246 (Stage 2)$\rightarrow$0.209 (Stage 3), a 19\% gain through multimodal expansion.

This improvement is particularly notable given three fundamental challenges in the field. First, the compression-dimension trade-off severely constrains 16$\times$16 models: VAVAE \citep{Yao2025ReconstructionVG} requires 32-dimensional latents to achieve 0.279 rFID, while Cosmos-CI16$\times$16 with 16 dimensions degrades to 0.959 rFID. Second, transformer architectures consistently underperform convolutional architectures (OmniTokenizer \citep{wang2024omnitokenizer} 26.74 PSNR vs. Hunyuan \citep{kong2024hunyuanvideo} 33.32 PSNR), explaining why most reconstruction tokenizers avoid transformers. Third, discrete tokenizers struggle with generalization -- UniTok~\citep{ma2025unitok} degrades from 0.362 rFID on ImageNet to 3.918 on COCO, while GigaTok~\citep{xiong2025gigatok} exhibits even larger gaps.

Our approach addresses all three challenges: achieving strong performance with 48-dimensional latents at 16$\times$16 compression, demonstrating transformer viability through adversarial-free training, and maintaining consistent quality across datasets (0.209 rFID on ImageNet, 2.026 rFID on COCO). These results suggest temporal dynamics from video and geometric understanding from 3D provide complementary signals for image reconstruction.

\input{table/imagenet_zeroshot}

\textbf{Semantic Understanding.} 
\cref{tab:main_image_zeroshot} evaluates zero-shot classification and retrieval against leading vision encoders. While understanding-only models like CLIP \citep{radford2021learning} and its variants \citep{xu2023metaclip, EVACLIP, Fang2023DataFN} optimize purely for semantic alignment, \atoken~need to balance understanding with reconstruction across three modalities.

Despite these constraints, \atoken~achieves 82.2\% ImageNet accuracy -- within 1.2\% of understanding-only SigLIP2 \citep{tschannen2025siglip} (83.4\%). This narrows the gap compared to previous unified attempts like UniTok (78.6\%) and VILA-U (78.0\%), while uniquely extending unified capabilities to video and 3D. Across our progressive training stages, accuracy remains stable (82.7\% → 82.3\% → 82.2\%), with only 0.5\% degradation as modalities are added. Discrete quantization also preserves full semantic performance, achieving 82.2\% accuracy.

\subsection{Video Tokenization}
\label{subsec:video}

We evaluate \atoken's video capabilities through reconstruction quality and semantic understanding benchmarks, demonstrating competitive performance while uniquely supporting both continuous and discrete representations across multiple modalities.

\textbf{Reconstruction Performance.} 
We evaluate video reconstruction on DAVIS \citep{PontTuset2017The2D} (1080p, 50 videos) and TokenBench \citep{Agarwal2025CosmosWF} (720p, 471 videos), reporting PSNR and SSIM for pixel quality, LPIPS for perceptual similarity, and rFVD for temporal consistency. All baselines were re-evaluated using official implementations with consistent protocols and spatial tiling for memory management. \atoken~employs temporal tiling with KV-caching, leveraging its native 2048$\times$2048 resolution support.

As shown in \cref{tab:video_rec}, \atoken-So/\textsc{c} achieves 33.11 PSNR on DAVIS and 36.07 PSNR on TokenBench, approaching specialized video-only models (Wan2.1 \citep{wan2025}: 33.50 and 36.11, Hunyuan \citep{kong2024hunyuanvideo}: 32.33 and 36.37). Notably, we demonstrate that transformers can match CNN performance when properly designed -- our method dramatically outperforms OmniTokenizer's transformer baseline (21.06 vs 33.11 PSNR on DAVIS) while adding native resolution support. Furthermore, our progressive training reveals cross-modal benefits: incorporating 3D in Stage 3 improves video reconstruction from 35.63 to 36.07 PSNR on TokenBench, indicating that geometric understanding may enhance temporal modeling.
For discrete tokenization, \atoken-So/\textsc{d} pioneers multimodal video support, achieving 29.75 PSNR on DAVIS -- surpassing Cosmos-0.1-DV (27.26) and dramatically outperforming OmniTokenizer (20.62), while maintaining reasonable perceptual quality (0.288 LPIPS) for downstream tasks.

\input{table/video_rec.tex}

\input{table/video_retrieval}

\textbf{Semantic Understanding.}
\cref{tab:video_zeroshot_results} evaluates zero-shot video-text retrieval on MSRVTT \citep{Xu2016MSRVTTAL} and MSVD \citep{chen2011collecting}. Following standard protocols \citep{Wang2022InternVideoGV, Luo2021CLIP4Clip}, we use frame embedding averaging with zero-padding. 
\atoken~achieves 40.2\% R@1 on MSRVTT and 53.5\% on MSVD, maintaining reasonable semantic alignment despite optimizing primarily for reconstruction across three modalities. We note that alternative pooling strategies without frame averaging yielded lower performance, likely due to the limited video-text pairs in our training data compared to dedicated video understanding models. While understanding-only models trained on large-scale video-text data achieve higher scores, our results validate that unified tokenization successfully balances reconstruction quality with semantic understanding.

\subsection{3D Tokenization.}
\label{subsec:3d}

We evaluate \atoken's 3D capabilities on Toys4k \citep{stojanov21cvpr} for reconstruction and semantic understanding. For reconstruction, \atoken-So/\textsc{c} achieves 28.28 PSNR and 0.062 LPIPS (\cref{tab:3d_reconstruction}), surpassing the specialized Trellis-SLAT \citep{xiang2024structured} baseline (26.97 PSNR, 0.054 LPIPS) despite jointly training across three modalities. This demonstrates that our unified 4D representation effectively captures geometric structure without requiring dedicated 3D architectures.

For semantic understanding, \atoken-So/\textsc{c} achieves 90.9\% zero-shot classification accuracy on Toys4k, validating that our approach maintains strong semantic representations for 3D objects alongside reconstruction capabilities. Combined with our image and video results, this confirms that all three modalities can coexist within a single tokenizer without significant trade-offs.

\subsection{Ablation Study}
\label{subsec:ablation}

\begin{figure*}[t]
    \centering
    \includegraphics[width=\textwidth]{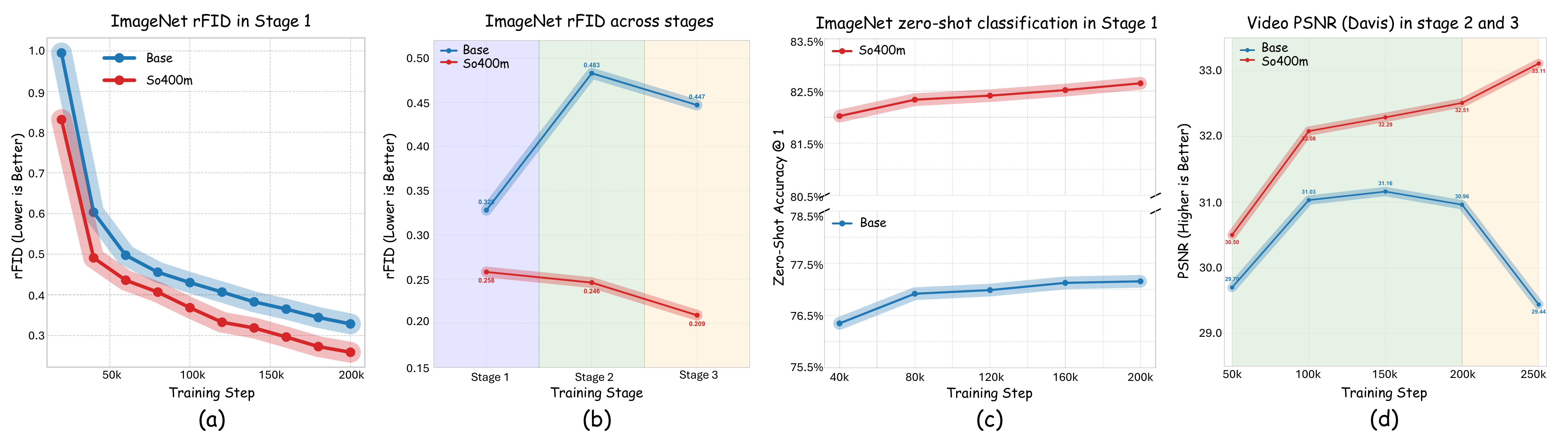}
    \caption{\small \textbf{Architectural scaling comparison: Base vs. So400m models.} 
    (a) ImageNet rFID during Stage 1 training. 
    (b) ImageNet rFID across training stages. 
    (c) ImageNet zero-shot classification accuracy in Stage 1. 
    (d) Video PSNR on DAVIS in Stages 2 and 3. 
    The So400m model maintains or improves performance across all stages, while the Base model shows significant degradation when extending beyond single-modality training, indicating that sufficient model capacity is critical for successful multimodal visual tokenization.}
    \label{fig:ablate_model_size}
\end{figure*}
\input{table/3d_rec}

\paragraph{Scaling Analysis.}

To investigate the scaling property of the visual tokenizer, we compare our So400m model with a smaller Base variant following identical training procedures. The Base model initializes from SigLIP-Base-patch16-naflex \citep{tschannen2025siglip}, comprising 12 transformer blocks with hidden dimension $d=768$ and 12 attention heads for both encoder and decoder, yielding approximately 192M parameters compared to So400m's 800M. 

As shown in \cref{fig:ablate_model_size}, both models achieve reasonable single-modal performance in Stage 1, with So400m outperforming Base (0.258 vs 0.323 rFID, 82.7\% vs 77.2\% accuracy).
However, the Base model suffers severe degradation when expanding to videos, with ImageNet rFID degrading 49\% (0.323$\rightarrow$0.483) and video PSNR declining across stages.
In contrast, So400m improves continuously -- ImageNet rFID enhances 19\% (0.258$\rightarrow$0.209) while video PSNR rises from 32.51 to 33.11. This scaling analysis reveals that multimodal tokenization has a capacity requirement: small models suffer from interference while large models benefit from cross-modal learning.

\begin{figure*}[t]
    \centering
    \includegraphics[width=\textwidth]{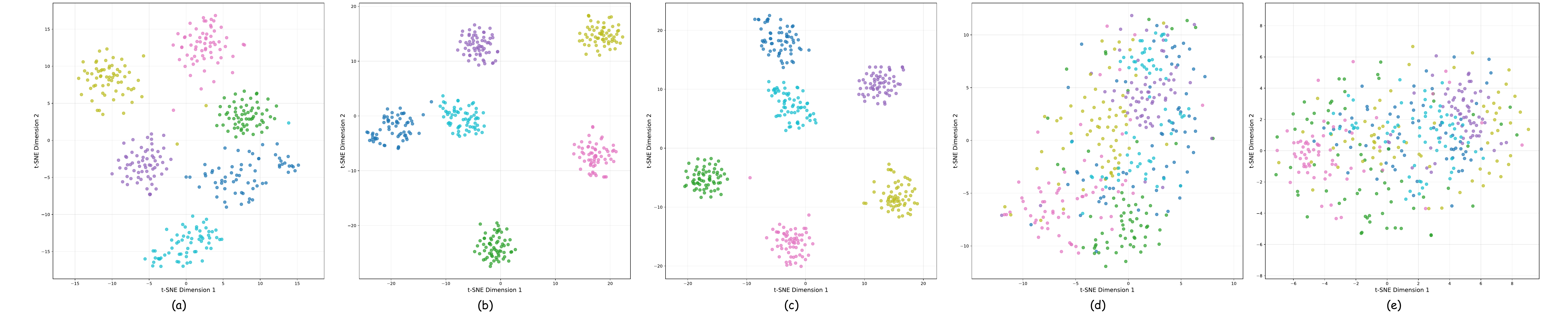}
    \caption{\small \textbf{Learned representations across training stages.}
    T-SNE visualizations of ImageNet class embeddings (colors indicate different classes). 
    (a) Stage 1: image-only training. 
    (b) Stage 2: with video. 
    (c) Stage 3: dense features before projection. 
    (d) Stage 3: projected 48-dim latents. 
    (e) Stage 4: before FSQ quantization. 
    Dense features (a-c) show clear semantic clustering, while dimensional reduction (d-e) leads to more mixed class distributions, suggesting a trade-off between compression and semantic separability.}
    \label{fig:visu_cluster}
    \vspace{-3mm}
\end{figure*}

\paragraph{Representation Structure Analysis.}

\cref{fig:visu_cluster} visualizes learned representations through T-SNE projections across training stages. Dense features (a-c) show clear semantic clustering with distinct ImageNet class separation. However, projection to 48-dimensional latents (d-e) results in more intermixed distributions, likely due to KL regularization without post-projection alignment loss. 

Despite this apparent mixing in T-SNE visualizations, the model maintains strong reconstruction and understanding performance, suggesting that semantic information may be encoded in ways not captured by 2D projections. This raises an interesting question: whether explicit semantic clustering in low-dimensional spaces -- as emphasized by methods like VAVAE \citep{Yao2025ReconstructionVG} -- is necessary for strong performance, or whether larger models can effectively leverage seemingly intermixed representations. Our results suggest the latter, though we leave detailed investigation of semantic preservation through aggressive dimensionality reduction for future work.

\paragraph{Reconstruction Visualization.}
Figures~\ref{fig:image_rec_compare}-\ref{fig:3d_rec_compare} provide qualitative comparisons of reconstruction quality across all three modalities. For images (\cref{fig:image_rec_compare}), \atoken~operates at a higher compression ratio (16$\times$) than most baselines yet achieves superior visual fidelity, particularly in preserving high-frequency details such as text clarity, fine textures, and complex patterns. The comparison reveals that methods optimized for lower compression ratios (e.g., SD-VAE and OmniTok at 8$\times$) struggle with text legibility and texture preservation, while \atoken~maintains sharp details. For video reconstruction (\cref{fig:video_rec_compare}), \atoken~demonstrates temporal consistency comparable to specialized video tokenizers like Wan2.2, with both continuous and discrete variants preserving motion smoothness across 720p sequences. The 3D reconstruction results (\cref{fig:3d_rec_compare}) highlight \atoken's advantage in color consistency. While Trellis-SLAT exhibits color shifts and artifacts, our unified training across modalities transfers color understanding from images and videos to improve 3D reconstruction.

\input{table/all_figure}

%% file: table/main_results.tex
\begin{table*}[t]
    \small
    \newcommand{\best}[1]{\bf{#1}}
    \caption{
        \small \textbf{Performance comparison of visual tokenizers across modalities.}
    We evaluate on ImageNet for image reconstruction and zero-shot classification, 
    TokenBench for video reconstruction with MSR-VTT for zero-shot retrieval, and Toys4k for 3D 
    reconstruction and classification. Methods are grouped by capability: reconstruction-only, 
    understanding-only, and unified approaches. Discrete tokenizers are indicated 
    with gray shading. $^\dagger$ OmniTokenizer does not work well on high-resolution videos where tiling is needed.
    }
    \label{tab:main_results}

    \centering
    \setlength\tabcolsep{4pt}
    \renewcommand{\arraystretch}{1.2} 
    \resizebox{\textwidth}{!}{
    \begin{tabular}{l ccc ccc ccc ccc}
        \toprule
        & \multirow[b]{2}{*}{\bf\renewcommand{\arraystretch}{1.0}\begin{tabular}{c}Comp.\\Ratio\end{tabular}} & \multirow[b]{2}{*}{\bf\renewcommand{\arraystretch}{1.0}\begin{tabular}{c}Latent\\Channels\end{tabular}} & 
        \multirow[b]{2}{*}{\bf\renewcommand{\arraystretch}{1.0}\begin{tabular}{c}Token\\Type\end{tabular}} & 
        \multicolumn{3}{c}{\bf Image} & \multicolumn{3}{c}{\bf Video} & \multicolumn{3}{c}{\bf 3D}  \\
        \cmidrule(lr){5-7} \cmidrule(lr){8-10} \cmidrule(lr){11-13}
        {\bf Method} & & & & {\bf PSNR$\uparrow$} & {\bf rFID$\downarrow$} & {\bf Acc.$\uparrow$} & {\bf PSNR$\uparrow$} & {\bf rFVD$\downarrow$} & {\bf R@1$\uparrow$} & {\bf PSNR$\uparrow$} & {\bf LPIPS$\downarrow$} & {\bf Acc.$\uparrow$}\\
        \midrule
        \multicolumn{13}{l}{\textit{\textbf{Reconstruction Only}}} \\
        \quad SD-VAE         & (1,~~~8,~~~8)   & 4  & VAE & 26.26          & 0.61         & -      & -     & -     & -      & -     & -     & - \\
        \quad FLUX.1 [dev]         & (1,~~~8,~~~8)   & 16  & VAE & 32.86          & \textbf{0.18}         & -      & -     & -     & -      & -     & -     & - \\

        % \quad VQ-GAN         & (1, 8, 8)   & 4  & VAE &          &         & -      & -     & -     & -      & -     & -     & - \\
        \quad Cosmos-0.1-CI8$\times$8  & (1,~~~8,~~~8)   & 16 & AE  & 32.25          & 1.03         & -      & -     & -     & -      & -     & -     & - \\
        \quad Qwen-Image & (1,~~~8,~~~8) & 16 & VAE & 32.18 & 1.46 & -      & -     & -     & -      & -     & -     & - \\
        \quad VA-VAE         & (1, 16, 16) & 32 & VAE & 27.70          & 0.28         & -      & -     & -     & -      & -     & -     & - \\
        \rowcolor{gray!15} \quad GigaTok-XL-XXL & (1, 16, 16) & 8 & VQ & 22.42 & 0.80 & -      & -     & -     & -      & -     & -     & - \\

        \quad Cosmos-0.1-CV8$\times$8  & (4,~~~8,~~~8) & 16 & AE  & 30.11             & 7.55             & -      & 34.33 & 8.34 & -      & -     & -     & - \\
        \quad OmniTokenizer$^\dagger$      & (4,~~~8,~~~8)   & 8 & VAE  & 26.74          & 1.02         & -      & 19.39  & 173.48  & -      & -     & -     & - \\ 
        \quad Hunyuan         & (4,~~~8,~~~8)   & 16 & VAE & \textbf{33.32}          & 0.67  & -      & 36.37 & 3.78 & -      & -     & -     & - \\
        \quad Wan2.1            & (4,~~~8,~~~8)   & 16 & VAE & 31.34          & 0.94         & -      & 36.11  & 3.21  & -      & -     & -     & - \\
        \quad Wan2.2          & (4, 16, 16)   & 48 & VAE & 31.25    & 0.75       & - & \textbf{36.39} & 3.19 & -      & -     & -     & - \\
        \rowcolor{gray!15} \quad OmniTokenizer$^\dagger$      & (4,~~~8,~~~8)   & 8 & VQ  & 24.69          & 1.41         & -      & 19.89    & 202.46     & -      & -     & -     & - \\ 
        \rowcolor{gray!15} \quad Cosmos-0.1-DV8$\times$8  & (4,~~~8,~~~8) & 6 & FSQ  & 26.34             & 7.86             & -      & 31.42 & 25.94 & -      & -     & -     & - \\
        \quad Trellis-SLAT   & -           & 8  & VAE & -             & -             & -      & -     & -     & -      & 26.97 & \textbf{0.054} & - \\
        
        \midrule
        \multicolumn{13}{l}{\textit{\textbf{Understanding Only}}} \\

        % \quad CLIP & (1, 14, 14) & - & - & -  & -  \\
        \quad VideoPrism-g & (1, 18, 18) & - & - & - & - & - & - & - & \textbf{52.7} & - & - & - \\
        \quad SigLIP2-So/16  & (1, 16, 16) & -  & -   & -             & -             & {83.4}   & -     & -     & 41.9      & -     & -     & - \\
        \quad PE$_\texttt{core}$L & (1, 14, 14) & - & - &  - & - & \textbf{83.5} & - & - & 50.3 & - & - & - \\
        \midrule
        \multicolumn{13}{l}{\textit{\textbf{Reconstruction \& Understanding}}} \\
        % \quad TokenFlow & (1, 16, 16) & -  & -   & -          & 1.37             & -      & -     & -     & -      & -     & -     & - \\
        \quad SeTok & - & 4096 & AE & - & 2.07 & 75.4 & - & - & - & - & - & -\\ 
        \rowcolor{gray!15} \quad VILA-U    & (1, 16, 16) & 16 & RQ  & 22.24          & 4.23             & 78.0   & -     & -     & -      & -     & -     & - \\
        \rowcolor{gray!15}  \quad UniTok    & (1, 16, 16) & 64 & MCQ & 25.34          & \textbf{0.36}             & 78.6   & -     & -     & -      & -     & -     & - \\
        \arrayrulecolor{gray!60}\cmidrule(r){1-13}
        \rowcolor{gray!15}  \quad \atokensod & (4, 16, 16) & 48 & FSQ & \textbf{27.00} & 0.38      &  82.2  & \textbf{33.12}   & \textbf{22.16}   &  40.3     &  \textbf{28.17} &  \textbf{0.063}  & \textbf{91.3} \\
        \quad \atokensoc & (4, 16, 16) & 48 & VAE & 29.72   & {0.21}         & 82.2   & 36.07     & \textbf{3.01}    & 40.2 & {28.28} & 0.062 & 90.9 \\
        \arrayrulecolor{black}\bottomrule
    \end{tabular}}
    \vspace{-3mm}
\end{table*}

%% file: table/image_rec_all.tex
\begin{table}[t]
    \small
    \caption{\small \textbf{Image reconstruction comparison on ImageNet and COCO.} 
We evaluate all methods using a unified protocol with official implementations 
to ensure fair comparison. All images are resized and center-cropped to 256$\times$256, 
with metrics computed using identical scripts. Note that our reproduced results may 
differ from original papers due to standardized evaluation settings, but provide 
consistent cross-model comparison.}
    \label{tab:main_image_rec}
    \newcommand{\best}[1]{\bf{#1}}
    \centering
      \resizebox{\columnwidth}{!}{
    \begin{tabular}{lccc cccc cccc}
        \toprule
         & \multirow[b]{2}{*}{\begin{tabular}{c}\textbf{Comp.}\\\textbf{Ratio}\end{tabular}} & \multirow[b]{2}{*}{\begin{tabular}{c}\textbf{Latent}\\\textbf{Size}\end{tabular}} & \multirow[b]{2}{*}{\begin{tabular}{c}\textbf{Token}\\\textbf{Type}\end{tabular}}  & \multicolumn{4}{c}{\textbf{ImageNet}} & \multicolumn{4}{c}{\textbf{COCO}} \\
        \cmidrule(lr){5-8} \cmidrule(lr){9-12}
        \textbf{Method} & & & & \textbf{PSNR}$\uparrow$ & \textbf{SSIM}$\uparrow$ & \textbf{LPIPS}$\downarrow$ & \textbf{rFID}$\downarrow$ & \textbf{PSNR}$\uparrow$ & \textbf{SSIM}$\uparrow$ & \textbf{LPIPS}$\downarrow$ & \textbf{rFID}$\downarrow$ \\
        \midrule
        \multicolumn{12}{l}{\cellcolor{gray!15}\textit{\textbf{Continuous Latent}}} \\
        \quad SD-VAE  & $(1,~~8,~~8)$ & 4 & VAE & 26.26 & 0.745 & 0.133 & 0.606 & 25.99 & 0.759 & 0.130 & 4.142 \\
        % Cosmos-0.1-CV8$\times$8 \cite{Agarwal2025CosmosWF} & $(4,~~8,~~8)$ &  & \\
        % Cosmos-0.1-CV16$\times$16 \cite{Agarwal2025CosmosWF} & $(8,16,16)$ &  &  \\
        \quad SD3-VAE  & $(1,~~8,~~8)$ & 16 & VAE & 31.29 & 0.886 & 0.059 & 0.201 & 31.18 & 0.894 & 0.056 & 1.671 \\
        \quad FLUX.1 [dev]  & $(1,~~8,~~8)$ & 16 & VAE & 32.86 & 0.917 & \textbf{0.044} & \textbf{0.176} & \textbf{32.73} & 0.923 & \textbf{0.041} & \textbf{1.343} \\
        \quad Qwen-Image  & $(1,~~8,~~8)$ & 16 & VAE & 32.18 & 0.899 & 0.053 & 1.459 & 32.01 & 0.908 & 0.050 & 4.618 \\
        \quad Cosmos-0.1-CI8$\times$8  & $(1,~~8,~~8)$ & 16 & AE & 32.25 & 0.902 & 0.064 & 1.031 & 32.08 & 0.909 & 0.061 & 3.844 \\
        \quad Cosmos-0.1-CI16$\times$16  & $(1, 16,16)$ & 16 & AE & 25.07 & 0.700 & 0.167 & 0.959 & 24.74 & 0.711 & 0.165 & 5.063 \\
        \quad VAVAE & $(1, 16,16)$ & 32 & VAE & 27.70 & 0.798 & 0.096 & 0.279 & 27.50 & 0.811 & 0.093 & 2.709 \\
        \quad OmniTokenizer  & $(4,~~8,~~8)$ & 8 & VAE & 26.74 & 0.824 & 0.101 & 1.023 & 26.44 & 0.833 & 0.099 & 4.687 \\
        \quad Hunyuan  & $(4,~~8,~~8)$ & 16 & VAE & \best{33.32} & \textbf{0.916} & 0.053 & 0.670 & 33.25 & \textbf{0.924} & 0.050 & 2.597 \\
        \quad Wan2.1  & $(4,~~8,~~8)$ & 16 & VAE & 31.34 & 0.886 & 0.058 & 0.945 & 31.19 & 0.895 & 0.055 & 3.449 \\
        \quad Wan2.2 & $(4,16,16)$ & 48 & VAE & 31.25 & 0.878 & 0.057 & 0.749 & 31.10 & 0.888 & 0.054 & 3.279 \\

        \arrayrulecolor{lightgray}\cmidrule(rl){1-12}
        \multicolumn{11}{l}{\atokensoc} \\
        \quad Stage 1 & $(1,16,16)$ & 32 & VAE & 28.77 & 0.814 & 0.099 & 0.258 & 28.66 & 0.829 & 0.096 & 2.336 \\
        \quad Stage 2 & $(4,16,16)$ & 48 & VAE & 29.55 & 0.845 & 0.087 & 0.246 & 29.49 & 0.858 & 0.083 & 2.180 \\
        \quad Stage 3 & $(4,16,16)$ & 48 & VAE & 29.72 & 0.848 & 0.085 & 0.209 & 29.67 & 0.861 &  0.081 &  2.026 \\

        \arrayrulecolor{black}\midrule
        \multicolumn{12}{l}{\cellcolor{gray!15}\textit{\textbf{Discrete Latent}}} \\
        \quad Cosmos-0.1-DI8$\times$8  & $(1,~~8,~~8)$ & 6 & FSQ & 25.87 & 0.750 & 0.155 & 0.867 & 25.54 & 0.760 & 0.153 & 5.016 \\
        \quad GigaTok-B-L  & $(1, 16,16)$ & 8 & VQ & 21.87 & 0.591 & 0.200 & 0.507 & 21.42 & 0.596 & 0.202 & 5.565 \\
        \quad GigaTok-XL-XXL  & $(1, 16,16)$ & 8 & VQ & 22.42 & 0.613 & 0.189 & 0.795 & 22.03 & 0.620 & 0.191 & 5.757 \\
        \quad Vila-U  & $(1, 16,16)$ & 16 & RQ & 22.24 & 0.612 & 0.228 & 4.231 & 21.89 & 0.620 & 0.227 & 10.997 \\
        \quad UniTok  & $(1, 16,16)$ & 64 & MCQ & 25.34 & 0.742 & 0.132 & \textbf{0.362} & 24.95 & 0.750 & 0.131 & 3.918 \\
        \quad OmniTokenizer  & $(4,~~8,~~8)$ & 8 & VQ & 24.69 & 0.771 & 0.138 & 1.411 & 24.31 & 0.779 & 0.137 & 6.292 \\
        \arrayrulecolor{lightgray}\cmidrule(rl){1-12}
        \quad \atokensod & $(4,16,16)$ & 48 & FSQ & \textbf{27.14} &\textbf{0.801} & \textbf{0.119} & 0.379 & \textbf{27.00} & \textbf{0.815} & \textbf{0.115} & \textbf{3.270}  \\
        \arrayrulecolor{black}\bottomrule
    \end{tabular}
    }
    \vspace{-3mm}
\end{table}

%% file: table/imagenet_zeroshot.tex
\begin{table}[t]
    \small
    \caption{\small \textbf{Image understanding comparison with semantic encoders.} We evaluate zero-shot classification on ImageNet, ImageNet-v2, and cross-modal retrieval on COCO and Flickr30k. \atoken~maintains competitive performance across all stages despite joint training on multiple modalities and tasks. }
    \label{tab:main_image_zeroshot}
    \centering
    \setlength{\tabcolsep}{0.5em}
    \resizebox{0.6\linewidth}{!}{
    \begin{tabular}{ccl@{\hspace{2.0em}}cccccc}
    \toprule
     &  & & \multicolumn{2}{c}{ImageNet-1k} & \multicolumn{2}{c}{COCO} & \multicolumn{2}{c}{Flickr} \\ \cmidrule(lr){4-5} \cmidrule(lr){6-7} \cmidrule(lr){8-9}
    Res. & Seq. & Model & val & v2 & T$\rightarrow$I & I$\rightarrow$T & T$\rightarrow$I & I$\rightarrow$T \\
    \midrule
    \multirow[c]{4}{*}{224} & \multirow[c]{4}{*}{196} 
        & CLIP      & 68.3 & 61.9 & 33.1 & 52.4 & 62.1 & 81.9 \\
     &  & MetaCLIP  & 72.4 & 65.1 & 48.9 & -- & 77.1 & -- \\
     &  & EVA-CLIP  & 74.7 & 67.0 & 42.2 & 58.7 & 71.2 & 85.7 \\
     &  & DFN       & 76.2 & 68.2 & 51.9 & -- & 77.3 & -- \\
    % \arrayrulecolor{lightgray}\hhline{|---------|} 
    \arrayrulecolor{black}\midrule
    \multirow[c]{8}{*}{256} & \multirow[c]{8}{*}{256} & SigLIP   & 80.8 & 74.1 & 49.4 & 68.6 & 80.0 & 92.1 \\
     &  & SigLIP 2 & \textbf{83.4} & \textbf{77.8} & \textbf{55.4} & \textbf{71.5} & \textbf{84.4} & \textbf{94.2} \\
    \arrayrulecolor{lightgray}\cmidrule(rl){3-9}
    & & \multicolumn{7}{l}{\atoken-So/\textsc{c}} \\
     &  & \quad Stage 1 & 82.7 & 76.7 & 54.1 & 70.4 & 81.3 & 93.1\\
     &  & \quad Stage 2 & 82.3 & 76.4 & 53.8 & 70.6 & 80.7 & 93.0\\
     &  & \quad Stage 3 & 82.2 & 76.1 & 53.7 & 70.5 & 80.5 & 93.2\\
     &  & \atoken-So/\textsc{d} & 82.2 & 76.2 & 53.8 & 70.1 & 80.9 & 93.5 \\
    \arrayrulecolor{black}\midrule
    % \arrayrulecolor{lightgray}\hhline{|---------|} 
    \multirow[c]{7}{*}{384} & \multirow[c]{7}{*}{576}  & SigLIP 2  & \textbf{84.1} & \textbf{78.4} & \textbf{56.0} & \textbf{71.2} & \textbf{85.3} & \textbf{95.9} \\
        \arrayrulecolor{lightgray}\cmidrule(rl){3-9}
    & & \multicolumn{7}{l}{\atoken-So/\textsc{c}} \\
     &  & \quad Stage 1 & 83.4 & 77.6 & 54.8 & 70.4 & 81.7 & 93.8\\
     &  & \quad Stage 2 & 82.9 & 77.1 & 54.7 & 71.1 & 81.9 & 93.9\\
     &  & \quad Stage 3 & 82.9 & 76.8 & 54.6 & 71.3 & 81.9 & 93.5\\
     &  & \atoken-So/\textsc{d} & 82.8 & 76.6 & 54.4 & 70.9 & 81.9 & 93.5\\
    \arrayrulecolor{black}\midrule
    % \arrayrulecolor{lightgray}\hhline{|---------|} 
    \multirow[c]{7}{*}{512} & \multirow[c]{7}{*}{1024} & SigLIP 2  & \textbf{84.3} & \textbf{79.1} & \textbf{56.0} & \textbf{71.3} & \textbf{85.5} & \textbf{95.4} \\
        \arrayrulecolor{lightgray}\cmidrule(rl){3-9}

    & & \multicolumn{7}{l}{\atoken-So/\textsc{c}} \\
     &  & \quad Stage 1 & 83.5 & 77.8 & 54.7 & 71.1 & 82.1 & 94.1\\
     &  & \quad Stage 2 & 83.1 & 77.3 & 54.7 & 71.3 & 82.2 & 93.6\\
     &  & \quad Stage 3 & 82.9 & 77.2 & 54.7 & 71.1 & 82.3 & 93.6\\
     &  & \atoken-So/\textsc{d} & 82.9 & 77.0 & 54.7 & 71.2 & 82.3 & 93.5 \\

    \arrayrulecolor{black}
    \bottomrule
    \end{tabular}
    }
    \vspace{-3mm}
\end{table}

%% file: table/video_rec.tex
\begin{table*}[!t]
\centering
\caption{\small \textbf{Video reconstruction comparison on high-resolution benchmarks.} 
We evaluate quality on DAVIS at 1080p and TokenBench at 720p. 
All methods are re-evaluated using official implementations with consistent protocols 
for fair comparison. \atoken~achieves competitive performance with specialized video-only 
tokenizers while uniquely supporting both continuous and discrete representations 
across modalities.
}
\label{tab:video_rec}

\resizebox{\textwidth}{!}{
\begin{tabular}{l ccc cccc cccc}
\toprule
 & \multirow[b]{2}{*}{\begin{tabular}{c}\textbf{Comp.}\\\textbf{Ratio}\end{tabular}} & \multirow[b]{2}{*}{\begin{tabular}{c}\textbf{Latent}\\\textbf{Size}\end{tabular}}  & \multirow[b]{2}{*}{\begin{tabular}{c}\textbf{Token}\\\textbf{Type}\end{tabular}}  & \multicolumn{4}{c}{\textbf{DAVIS}} & \multicolumn{4}{c}{\textbf{TokenBench}} \\
\cmidrule(lr){5-8} \cmidrule(lr){9-12}
\textbf{Tokenizer} & & & & \textbf{PSNR}$\uparrow$ & \textbf{SSIM}$\uparrow$ & \textbf{LPIPS}$\downarrow$ & \textbf{rFVD}$\downarrow$ & \textbf{PSNR}$\uparrow$ & \textbf{SSIM}$\uparrow$ & \textbf{LPIPS}$\downarrow$ & \textbf{rFVD}$\downarrow$ \\
\midrule
\multicolumn{12}{l}{\cellcolor{gray!15}\textit{\textbf{Continuous Latent}}} \\
\quad Cosmos-0.1-CV4$\times$8$\times$8 & (4,~~~8,~~~8) & 16 & AE & 32.25 & 0.894 & 0.219 & 19.15 & 34.33 & 0.924 & 0.155 & 8.34 \\
\quad OmniTokenizer & (4,~~~8,~~~8) & 8 & VAE & 21.06 & 0.800 & 0.315 & 206.34 & 19.39 & 0.782 & 0.275 & 173.48 \\
\quad Hunyuan & (4,~~~8,~~~8) & 16 & VAE & 32.33  & \textbf{0.907} & 0.194 &  22.94 & 36.37 & \textbf{0.944} & 0.129 &  3.78 \\
\quad Wan2.1 & (4,~~~8,~~~8)  & 16 & VAE & \textbf{33.50}  & 0.884 & \textbf{0.164} & 17.75 & 36.11 & 0.940 & 0.128 & 3.21 \\
\quad Wan2.2 & (4, 16, 16) & 48 & VAE & 33.06  & \textbf{0.907}       &  0.184      & 12.65      &  \textbf{36.39}  & 0.942   &  \textbf{0.126}    & 3.19 \\
\arrayrulecolor{lightgray}\cmidrule(rl){1-12}
\multicolumn{7}{l}{\atoken-So/\textsc{c}} \\
        \quad Stage 2 & (4, 16, 16)  & 48 & VAE & 32.29 & 0.902 & 0.196 & 13.50 & 35.63  & 0.937 & 0.139  & 3.63 \\
        \quad Stage 3 & (4, 16, 16)  & 48 & VAE & 33.11 & \textbf{0.907} & 0.189 & \textbf{10.76} & 36.07 & 0.940 & 0.135 & \textbf{3.01} \\

\arrayrulecolor{black}\midrule
\multicolumn{12}{l}{\cellcolor{gray!15}\textit{\textbf{Discrete Latent}}} \\

\quad OmniTokenizer & (4,~~~8,~~~8)  & 8 & VQ & 20.62 & 0.770 & 0.346 & 240.20 & 19.89 & 0.787 & 0.293 & 202.46\\
\quad Cosmos-0.1-DV4$\times$8$\times$8 & (4,~~~8,~~~8)  & 6 & FSQ & 27.26 & 0.798 & 0.310 & 110.33 & 31.20 & 0.892 & \textbf{0.190} & 25.94\\
\arrayrulecolor{lightgray}\cmidrule(rl){1-12}
\quad \atoken-So/\textsc{d} & (4, 16, 16)  & 48 & FSQ & \textbf{29.75} & \textbf{0.846} & \textbf{0.288} & \textbf{41.42} &  \textbf{33.12} & \textbf{0.913} &  0.193 & \textbf{22.16}\\
\arrayrulecolor{black}\bottomrule
\end{tabular}
}
\end{table*}

%% file: table/video_retrieval.tex
\begin{table*}[!t]
\centering

\caption{\small \textbf{Zero-shot video-text retrieval on MSRVTT and MSVD.} 
We compare \atoken~against understanding-focused encoders on standard video retrieval benchmarks. Despite optimizing for both reconstruction and understanding across three modalities, \atoken~maintains reasonable retrieval performance.}

\label{tab:video_zeroshot_results}
\setlength\tabcolsep{7pt}
\resizebox{\textwidth}{!}{
\begin{tabular}{l@{\hspace{2.0em}} c@{\hspace{2.0em}} cccccc cccccc}
\toprule
 & & \multicolumn{6}{c}{\textbf{MSRVTT (1K-A)}} & \multicolumn{6}{c}{\textbf{MSVD}} \\
\cmidrule(lr){3-8} \cmidrule(lr){9-14}
Methods & Res. & \multicolumn{3}{c}{Text $\rightarrow$ Video} & \multicolumn{3}{c}{Video $\rightarrow$ Text}  & \multicolumn{3}{c}{Text $\rightarrow$ Video} & \multicolumn{3}{c}{Video $\rightarrow$ Text} \\
\cmidrule(lr){3-5} \cmidrule(lr){6-8} \cmidrule(lr){9-11} \cmidrule(lr){12-14}
& & R@1 & R@5 & R@10 & R@1 & R@5 & R@10 & R@1 & R@5 & R@10 & R@1 & R@5 & R@10\\
\midrule
CLIP-ViT-B/32 & 224 & 31.2 & 53.7 & 63.3 & 26.4 & 49.9 & 61.7 & 36.4 & 63.3 & 73.1 & 57.8	& 84.1 & 90.7 \\
SigLIP2-So400m & 256 & 41.9 & 66.3 &  75.7 & 32.4 & 55.4 & 65.9 & \textbf{55.5} & \textbf{81.2} & 87.8 & 72.7 & 91.7 & 96.1\\
VideoPrism-g & 288 & \textbf{52.7} & \textbf{77.2} & - & \textbf{51.7} & \textbf{75.2} & - & - & - & - & - & - & - \\
PE-Core-B16 & 224 & 45.8 & 70.1 & 78.1 & 45.5 & 70.9 & 80.0 & 48.7 & 75.5 & 84.1 & 79.1 & 96.7 & 98.8\\
PE-Core-L14 & 336 & 49.1 & 73.3 & \textbf{81.6} & 50.9 & 74.4 & \textbf{82.7} & 54.4 & \textbf{81.2} & \textbf{88.4} & \textbf{82.5} & \textbf{98.2} & \textbf{99.4} \\
\midrule
\multicolumn{13}{l}{\atoken-So/\textsc{c}--224} \\
\quad Stage 1 & 224 & 40.8 & 65.3 & 75.2 & 31.0 & 55.0 & 63.7 & 53.9 & 79.9 & 87.3 & 72.4 & 93.0 & 95.4\\
\quad Stage 2 & 224 & 40.1 & 64.9 & 75.2 & 30.9 & 53.7 & 64.0 &  53.4 & 79.6 & 87.1 & 71.6 & 91.9 & 95.5 \\
\quad Stage 3 & 224 & 40.2 & 64.9 & 75.2 & 30.5 & 53.1 & 63.2 & 53.5 & 79.5 & 87.1 & 72.4 & 91.6 & 95.4 \\
\atokensod & 224 & 40.3 & 65.0 & 74.6 & 30.3 & 51.8 & 61.7 & 53.8 & 79.7 & 87.2 & 71.5 & 91.8 & 95.2\\

\bottomrule
\end{tabular}
}
\vspace{-3mm}
\end{table*}

%% file: table/3d_rec.tex
\begin{table}[t]
    \centering
    \caption{\small \textbf{3D reconstruction comparison on Toys4k.} 
    We average metrics across rendered multi-view images. 
    \atoken~achieves comparable performance to specialized Trellis-SLAT despite jointly optimizing 
    for three modalities, demonstrating unified training maintains strong 
    3D capabilities.}
    \label{tab:3d_reconstruction}
    \resizebox{0.45\columnwidth}{!}{
    \begin{tabular}{l cccc}
        \toprule
        \textbf{Method} & \textbf{PSNR$\uparrow$} & \textbf{SSIM$\uparrow$} & \textbf{LPIPS$\downarrow$} \\
        \midrule
        \multicolumn{5}{l}{\cellcolor{gray!15}\textit{Specialized 3D Tokenizer}} \\
        \quad Trellis-SLAT & 26.97 & 0.943 & \textbf{0.054} \\
        \arrayrulecolor{lightgray}\cmidrule(rl){1-5}
        \multicolumn{5}{l}{\cellcolor{gray!15}\textit{Our Unified Tokenizer (\atoken)}} \\
        \quad \atoken~-So/\textsc{c} & \textbf{28.28} & \textbf{0.951} & 0.062 \\
        \quad \atoken~-So/\textsc{d} & 28.17 & \textbf{0.951} & 0.063 \\
        \arrayrulecolor{black}  
        \bottomrule
    \end{tabular}
    }
    \vspace{-3mm}
\end{table}

%% file: table/all_figure.tex
\begin{figure}[!htb]
    \small
    \centering
    
    \includegraphics[width=\textwidth]{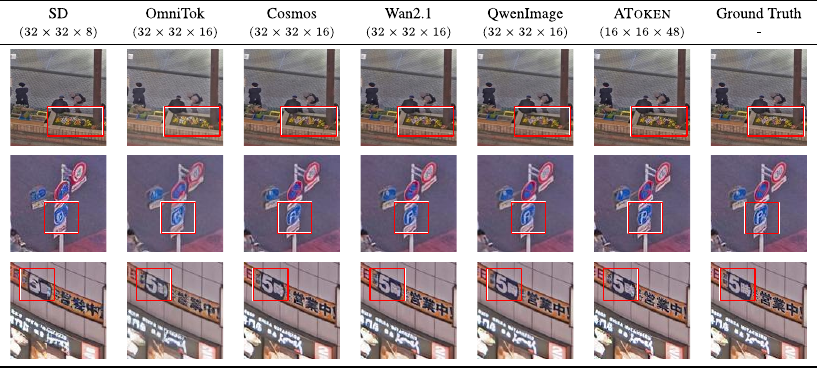}
    \caption{\small \textbf{Qualitative comparison of image reconstruction performance across different tokenization methods.} The latent shape for a $256 \times 256$ image patch is shown under each method name. Despite operating at higher compression ratios, \atoken~demonstrates superior reconstruction quality, particularly excelling in preserving high-frequency textures, fine details, and complex text elements.}
    \label{fig:image_rec_compare}
    
    \vspace{1em} % Space between subfigures
    
    % Second: Video figure
    \includegraphics[width=1.0\linewidth]{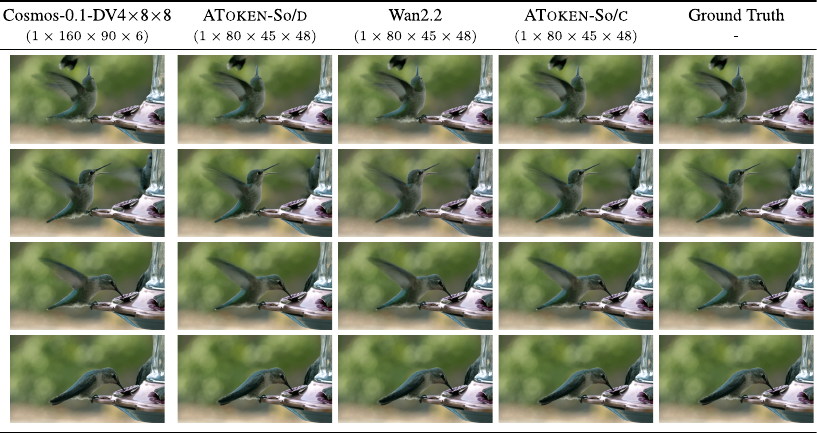}
    \caption{\small \textbf{Qualitative comparison of video reconstruction performance on 720p video sequences.} The latent shape for each video tokenization method is indicated under the method name. \atoken~achieves comparable quality to specialized video-only methods while uniquely supporting both continuous and discrete representations in a unified framework.}

    \label{fig:video_rec_compare}
    
    \vspace{1em} % Space between subfigures
    
    % Third: 3D reconstruction
    \includegraphics[width=1.0\textwidth]{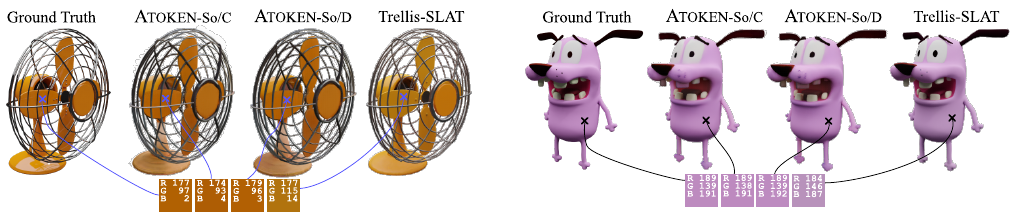}
    \caption{\small \textbf{3D Reconstruction Visualization on Toys4k.} \atoken's improved color consistency results in a higher PSNR compared to specialized 3D tokenizer Trellis-SLAT. }

    \label{fig:3d_rec_compare}
    
\end{figure}

%% file: sec/5_application.tex
\section{Downstream Results}
\label{sec:downstream}

Having established \atoken's unified tokenization capabilities across modalities, we evaluate its effectiveness in diverse downstream applications. We assess both understanding tasks through multimodal LLMs (\cref{subsec:mllm}) and generation tasks across images, videos, and 3D assets (Sections~\ref{subsec:image_gen_continous}--\ref{subsec:text_to_3d}). These experiments demonstrate that a single unified tokenizer can serve as the foundation for multimodal AI systems without compromising task-specific performance.

\subsection{Multimodal LLMs}
\label{subsec:mllm}

To validate \atoken's effectiveness for vision-language understanding, we integrate it into SlowFast-LLaVA-1.5~\citep{xu2025slowfast}, replacing the Oryx-ViT~\citep{liu2024oryx} vision encoder with \atoken-So/C while keeping all other settings identical. To assess generalization, the \atoken~parameters are frozen during training, with only the SlowFast projector and LLM updated. We evaluate using the \texttt{lmms-eval}~\citep{zhang2024lmmsevalrealitycheckevaluation} toolkit and report official metrics without output filtering.

\paragraph{Image Understanding.}

\cref{tab:image_understanding_results} shows the image understanding results on 7 standard benchmarks, including RW-QA\footnote{https://huggingface.co/datasets/xai-org/RealworldQA}, AI2D~\citep{kembhavi2016diagram}, SQA~\citep{lu2022learn}, and MMMU~\citep{yue2023mmmu}, and MathVISTA~\citep{lu2024mathvista} for general image QA, as well as OCRBench~\citep{Liu_2024_ocrbench} and TextVQA~\citep{singh2019towards} for text and document understanding.
To position our models relative to state-of-the-art methods, we compare it against LLaVA-OV~\citep{li2024llava}, MM1.5~\citep{zhang2024mm1}, Molmo~\citep{deitke2024molmo}, BLIP3~\citep{xue2024xgen}, Phi-3.5-V~\citep{abdin2024phi}, InternVL2.5~\citep{zhang2024internlm}, and Qwen2-VL~\citep{wang2024qwen2}.

Here we highlight some key observations.
\textit{First,} compared to Oryx-ViT, a specific vision encoder for multimodal understanding,
SlowFast-LLaVA-1.5 with \atoken~as vision encoder shows overall better performance on image understanding across different model scales.
Specifically, \cref{tab:image_understanding_results} shows that SlowFast-LLaVA-1.5-7B with \atoken~outperforms Oryx-ViT under the same MLLM by 1.3\% on RW-QA, 1.0\% on SQA, and 1.3\% on TextVQA.
\textit{Second}, \atoken~shows strong generalization ability across different tasks and model scales.
For reference, using \atoken, SlowFast-LLaVA-1.5-3B achieves superior results on almost all benchmarks.
On RW-QA and AI2D, \atoken~outperforms Oryx-ViT across the 1B, 3B, and 7B scales and achieves very competitive performance.

\input{table/image_und}
\input{table/video_und}

\paragraph{Video Understanding.}

The video understanding results are summarized in \cref{tab:video_understanding_results}, covering a range of video tasks.
Video-MME~\citep{fu2024video}, PercepTest~\citep{patraucean2023perception}, and NExT-QA~\citep{xiao2021next} assess general video QA, whereas LongVideoBench~\citep{wu2025longvideobench}, MLVU~\citep{zhou2024mlvu}, and LVBench~\citep{wang2024lvbench} focus on temporal understanding on long-range context.
We compared with both video specialist models, such as Apollo~\citep{zohar2024apollo}, LLaVA-Video~\citep{llava178k}, and LinVT~\citep{gao2024linvt}, and unified image-video MLLMs, such as Oryx1.5~\citep{liu2024oryx}, InternVL2.5~\citep{zhang2024internlm}, and Qwen2VL~\citep{wang2024qwen2}.

We outline several key observations.
\textit{First}, \atoken~excels at smaller model scales. For reference, SlowFast-LLaVA-1.5-1.5B with \atoken~achieves
state-of-the-art performance on almost all benchmarks (\eg, outperforming Oryx-ViT by 0.8\% on LongVideoBench and 1.4\% on LVBench).
\textit{Second}, \atoken provides more performance gain on general video QA benchmarks.
Specifically, it achieves state-of-the-art results on VideoMME (\eg, 64.5\% with 7B LLM) and PercepTest (\eg, 70.3\% with 7B LLM) across scales.
\textit{Third}, we note the strong performance of Oryx-ViT on long-form video understanding, particularly on MLVU. We hypothesize that this advantage arises because (\textit{i}) Oryx-ViT was specifically designed for video understanding in LLMs and (\textit{ii}) it was trained on long-video retrieval tasks. Future work to address this gap includes incorporating more long videos into our training data to strengthen temporal modeling over long-range context.

\subsection{Image Generation with Continuous Tokens}
\label{subsec:image_gen_continous}

To evaluate \atoken's generative capabilities with continuous tokens, we assess class-conditional ImageNet generation using the Lightning-DiT~\citep{Yao2025ReconstructionVG} framework. We compare against both general diffusion methods (DiT~\citep{Peebles2022DiT}, SiT~\citep{ma2024sit}) and reconstruction-specialized approaches (REPA~\citep{yu2024representation}, VAVAE~\citep{Yao2025ReconstructionVG}). For fair comparison with VAVAE -- a strong baseline optimized specifically for image reconstruction through DINOv2 alignment -- we use identical training code, only adapting the input layer for \atoken's 48-dimensional latents (vs. 32 for VAVAE).

We follow standard CFG protocol by applying guidance across all latent channels, using scale 1.65 for our 48-channel models (vs. 1.5 for 32-channel models), consistent with Lightning-DiT findings that wider latents benefit from stronger guidance. Note that VAVAE applies CFG only to the first three channels as reported in their work.

As shown in \cref{tab:main_image_t2i}, \atoken-So/\textsc{c} Stage 3 achieves 1.56 gFID, competitive with specialized tokenizers despite optimizing for multiple modalities and tasks simultaneously. While VAVAE achieves 1.35 gFID through image-specific optimization and REPA reaches 1.42 through specialized reconstruction alignment, \atoken~demonstrates that unified tokenization can approach specialized performance without sacrificing versatility. Notably, our Base model shows consistent performance across stages (1.44$\rightarrow$1.54$\rightarrow$1.58 gFID), while the So model improves from Stage 2 to Stage 3 (1.88$\rightarrow$1.56), suggesting that multimodal training can enhance generation quality.

\input{table/imagenet_gen.tex}
\input{table/imagenet_gen_discrete.tex}

\subsection{Image Generation with Discrete Tokens}
\label{subsec:image_gen_discrete}

To evaluate \atokensod's generative capabilities, we integrate it into the TokenBridge~\citep{wang2025bridging} autoregressive framework, replacing only the tokenizer while maintaining all other settings. The key architectural difference lies in token representation: TokenBridge uses 16 dimensions with 8-level vocabularies, while \atoken-So/D uses 8 dimensions with 4096-level vocabularies—a more challenging configuration that requires modeling larger discrete spaces. Additionally, TokenBridge employs FFT-based dimension ordering to generate low-frequency structure first, whereas our model uses sequential generation. Following TokenBridge's evaluation protocol, we sample 50,000 images with CFG scale 3.1.
  
As shown in \cref{tab:main_image_t2i_discrete}, \atokensod~achieves a gFID of 2.23, demonstrating competitive performance against specialized discrete tokenizers including LFQ~\citep{Yu2023LanguageMB}, TikTok-L~\citep{yu2024image}, VQGAN~\citep{esser2020taming}, UniTok~\citep{unitok}, and TokenBridge~\citep{wang2025bridging}. While TokenBridge achieves lower gFID (1.76), this gap is expected given our larger vocabulary size (4096 vs. 8) and lack of frequency-based ordering optimization. Notably, we outperform UniTok (2.51 gFID), the only other unified visual tokenizer, demonstrating that multimodal capabilities need not compromise generation quality.

\subsection{Text to Video Generation}
\label{subsec:text_to_video}

To assess the text-to-video (T2V) capabilities of the \atoken-So/\textsc{c} tokenizers, we integrate them into a video generation model. Our model is built upon the MMDiT backbone~\citep{Esser2024ScalingRF} and incorporates design elements from recent video architectures~\citep{wan2025,kong2024hunyuanvideo,peng2025open}. Due to computational constraints, we conduct experiments with smaller models and limited training data, maintaining consistent settings across all tokenizers for fair comparison. Following a standard two-stage training approach, we first pretrain the model from scratch on text-to-image (T2I) tasks with each tokenizer. We then adapt this image model for video generation, enabling evaluation on both T2I and T2V benchmarks. To provide a fair and efficient basis for comparing tokenizers, all training is conducted at low resolutions, using 256$\times$256 for images and 192$\times$336 for videos.

For T2I evaluation, we report CLIP-Score~\citep{hessel2021clipscore}, Pick-Score~\citep{kirstain2023pick}, and GenEval~\citep{ghosh2023geneval}. For T2V tasks, we evaluate performance using the VBench benchmark~\citep{huang2024vbench}. We compare our results against state-of-the-art video tokenizers, namely Cosmos~\citep{Agarwal2025CosmosWF}, Hunyuan~\citep{kong2024hunyuanvideo}, and Wan~\citep{wan2025}. To ensure a fair comparison, we normalize the effective token budget for video generation across all tokenizers by adjusting the patch size. For example, we use a patch size of 2$\times$2 for 8$\times$8 spatial compression and 1$\times$1 for 16$\times$16 compression. Additionally, for T2V generation, we adjust the classifier free guidance (CFG) scale to account for differences in channel size, using a scale of 9.0 for a channel size of 48 and 4.5 for a channel size of 16.

\input{table/video_gen.tex}

As shown in \cref{tab:video_generation_results}, our \atoken-So/\textsc{c} tokenizers achieve results comparable to specialized video-optimized tokenizers across all metrics, outperforming Cosmos and matching the performance of Hunyuan and Wan, even though ours are designed for a broader range of tasks.

\subsection{Image to 3D Synthesis}
\label{subsec:text_to_3d}
To validate the utility of our learned discrete tokens for downstream generative tasks, we train an image-to-3D synthesis model. Following the methodology of Trellis-SLAT~\citep{xiang2024structured}, we adopt their diffusion model architecture and training regimen. We replace their original 3D tokens with the tokens generated by our \atoken-So/\textsc{c}. For a fair comparison, all inference hyperparameters, such as the number of diffusion steps and classifier-free guidance scale, are kept identical to those reported in the original work.

\begin{figure}[!htb]
    \small
    \centering

    \includegraphics[width=0.98\linewidth]{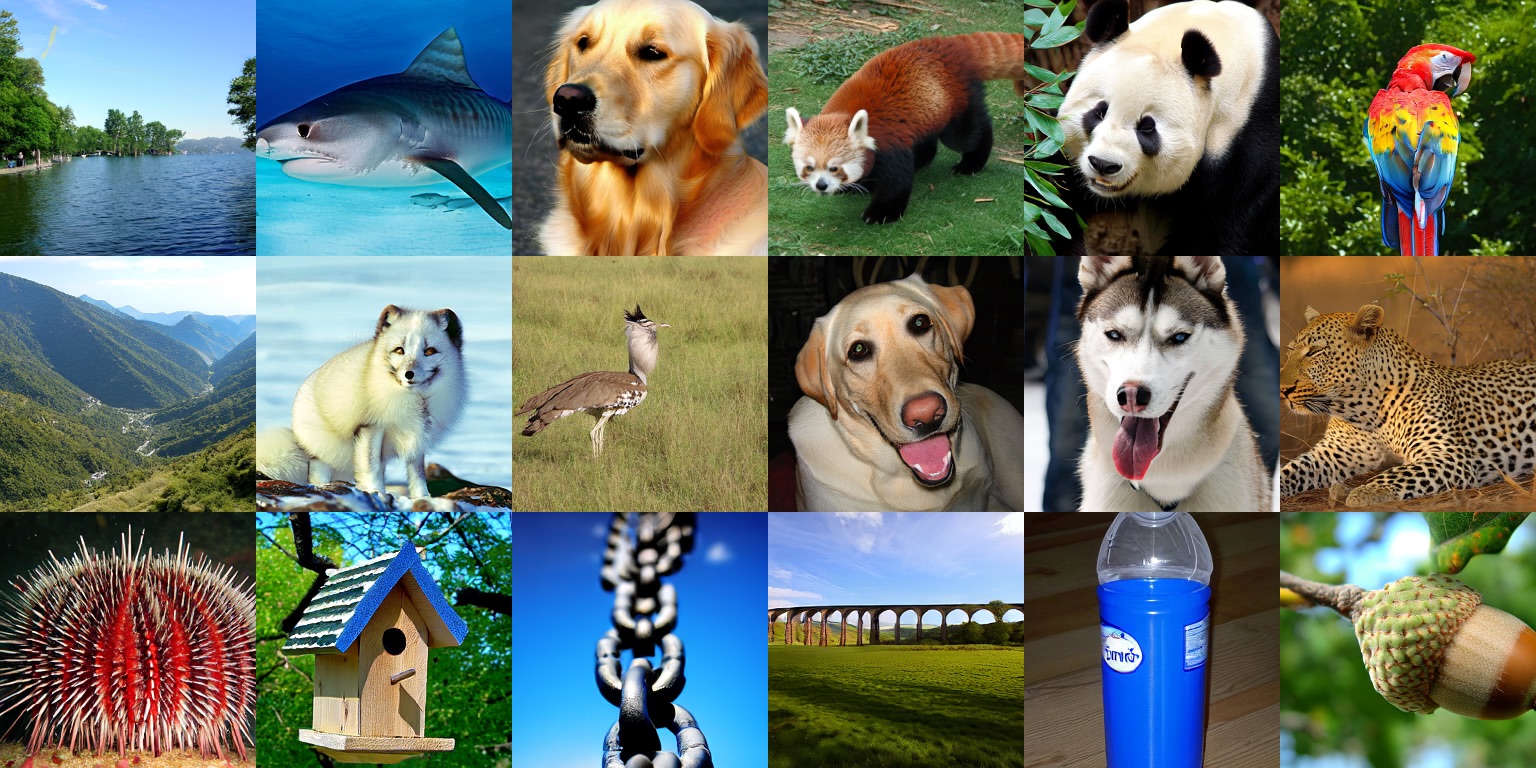}
    \caption{\small \textbf{ImageNet Generation Samples Using Continuous Token.} Images are generated with LightiningDiT \citep{Yao2025ReconstructionVG} and \atoken-So/\textsc{c}.}
    \label{fig:enter-label}
    
    \vspace{1em} % Space between subfigures
    
    % Second: Video figure
    \includegraphics[width=0.98\linewidth]{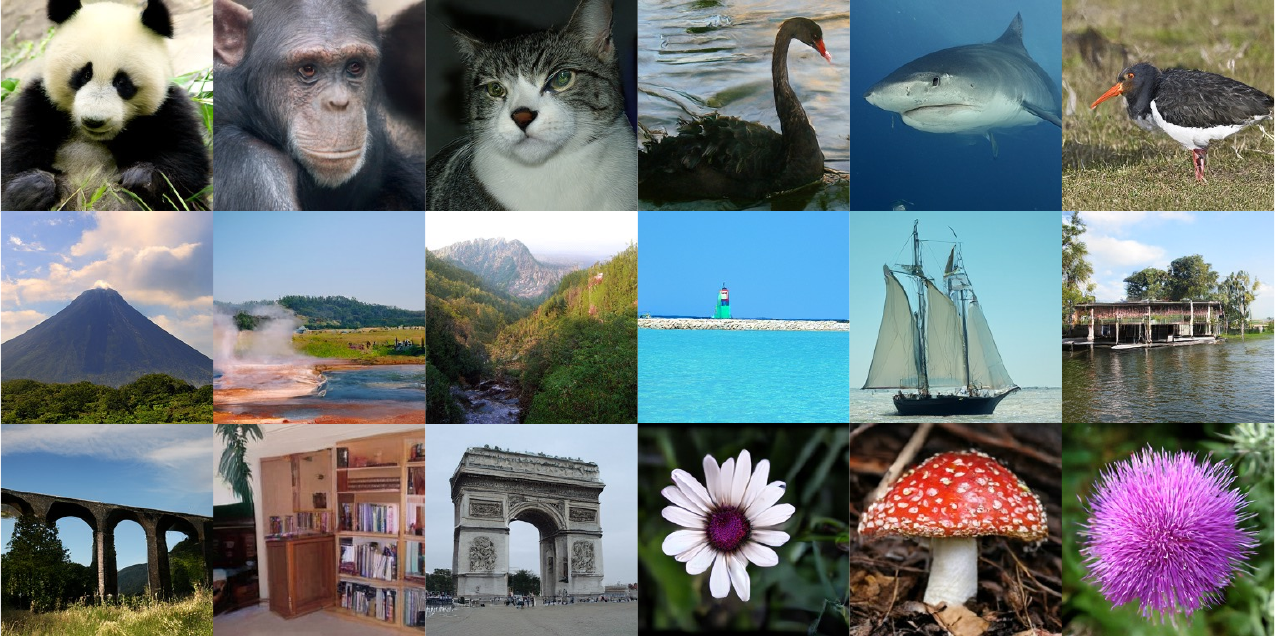}
    \caption{\small \textbf{ImageNet Generation Samples Using Discrete Token.} Images are generated with TokenBridge-L \citep{wang2025bridging} and  \atoken-So/\textsc{d}.}
    \label{fig:image_gen_discrete}
    
    \vspace{1em} % Space between subfigures
    
    % Third: 3D gen
    \includegraphics[width=0.98\textwidth]{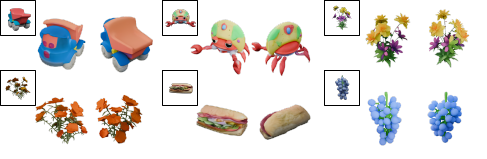}
    \caption{\small \textbf{Image-to-3D Generation Visualization on Toys4k.}}
    \label{fig:3d_gen}
    
\end{figure}

As shown in \cref{fig:3d_gen}, our approach successfully generates 3D assets from single conditioning images, demonstrating that our tokens are suitable for complex generative modeling. However, we observe that the performance does not yet match the fidelity of the original Trellis-SLAT model. Specifically, while our tokenizer demonstrates excellent reconstruction capabilities that preserve color and structure (as in \cref{fig:3d_rec_compare}), the generative model sometimes struggles to maintain this consistency. The generated assets do not always adhere strictly to the color and style of the input image.

We hypothesize that this discrepancy arises from the significantly larger latent channel dimension of our tokenizer. \atoken-So/\textsc{c} uses 48 latent channels to accommodate rich multimodal information, a substantial increase from the 8 channels used in Trellis-SLAT. A diffusion model operating in this higher-dimensional space likely requires further optimization of training and inference hyperparameters (e.g., conditioning strength, diffusion schedule) to leverage the conditioning signal fully. We leave the exploration of these optimizations as a promising direction for future work.

%% file: table/image_und.tex
\begin{table*}[!t]
\centering
\caption{\small 
\textbf{Image understanding comparison across multimodal LLMs.} 
Evaluation of SlowFast-LLaVA-1.5 with frozen \atoken-So/\textsc{c} vision encoder versus Oryx-ViT and other state-of-the-art MLLMs. Results shown for 7 benchmarks (general QA and text-rich understanding) across 1B, 3B, and 7B model scales.
}

\label{tab:image_understanding_results}
\resizebox{\textwidth}{!}{
\begin{tabular}{l cc cccccc cc}
\toprule
 & \multirow[b]{3}{*}{\begin{tabular}{c}\textbf{Vision}\\\textbf{Encoder}\end{tabular}} & \multirow[b]{3}{*}{\begin{tabular}{c}\textbf{\# Input}\\\textbf{Pixels}\end{tabular}} & \multicolumn{5}{c}{\textbf{General \& Knowledge}} & \multicolumn{2}{c}{\textbf{TextRich}} \\
\cmidrule(lr){4-8} \cmidrule(lr){9-10}
& & & \textbf{RW-QA} & \textbf{AI2D} & \textbf{SQA} & \textbf{MMMU} & \textbf{MathV} & \textbf{OCRBench} & \textbf{TextVQA} \\
\textbf{Multimodal LLM} & & & \textbf{(test)} & \textbf{(test)} & \textbf{(test)} & \textbf{(val)} & \textbf{(testmini)} & \textbf{(test)} & \textbf{(val)} \\
\midrule
\rowcolor{gray!15} \multicolumn{10}{l}{\textit{\textbf{1B Model Comparison}}} \\
        \quad LLaVA-OV-0.5B & SigLIP & 5.31M & 55.6 & 57.1 & 67.2 & 31.4 & 34.8 & - & - \\
        \quad MM1.5-1B & CLIP & 4.52M & 53.3 & 59.3 & 82.1 & 35.8 & 37.2 & 60.5 & 72.5 \\
        \quad MolmoE-1B & MetaCLIP & 4.10M & 60.4 & 86.4 & - & 34.9 & 34.0 & - & 78.8 \\
        \quad SlowFast-LLaVA-1.5-1B & Oryx-ViT & 2.36M & 59.2 & 72.8 & 87.7 & 40.5 & 51.0 & 70.0 & 71.3 \\
        \arrayrulecolor{lightgray}\cmidrule(rl){1-10}
        \quad SlowFast-LLaVA-1.5-1B &  \atoken-So/\textsc{c} & 2.36M & 60.1 & 74.2 & 88.7 & 40.6 & 52.5 & 67.6 & 72.5 \\
\arrayrulecolor{black}\midrule
\rowcolor{gray!15} \multicolumn{10}{l}{\textit{\textbf{3B Model Comparison}}} \\
        \quad BLIP3-4B & SigLIP & - & 60.5 & - & 88.3 & 41.1 & 39.6 & - & 71.0\\
        \quad MM1.5-3B & CLIP & 4.52M & 56.9 & 65.7 & 85.8 & 37.1 & 44.4 & 65.7 & 76.5 \\
        \quad Phi-3.5-V-4B & CLIP & - & - & 78.1 & 91.3 & 43.0 & 43.9 & - & 72.0 \\
        \quad SlowFast-LLaVA-1.5-3B & Oryx-ViT & 2.36M & 63.4 & 77.0 & 90.3 & 44.7 & 58.6 & 73.4 & 73.0 \\
        \arrayrulecolor{lightgray}\cmidrule(rl){1-10}
        \quad SlowFast-LLaVA-1.5-3B & \atoken-So/\textsc{c} & 2.36M & 64.3 & 79.1 & 89.7 & 45.7 & 58.4 & 73.3 & 72.8 \\
\arrayrulecolor{black}\midrule
\rowcolor{gray!15} \multicolumn{10}{l}{\textit{\textbf{7B Model Comparison}}} \\
        \quad LLaVA-OV-7B & SigLIP & 5.31M & 66.3 & 81.4 & 96.0 & 48.8 & 63.2 & - & - \\
        \quad MM1.5-7B & CLIP & 4.52M & 62.5 & 72.2 & 89.6 & 41.8 & 47.6 & 63.5 & 76.5 \\
        \quad Oryx1.5-7B & Oryx-ViT & 2.36M & - & 79.7 & - & 47.1 & - & 71.3 & 75.7 \\
        \quad InternVL2.5-8B & InternViT & 9.63M & 70.1 & 84.5 & - & 56.0 & 64.4 & - & 79.1 \\
        \quad Qwen2-VL-7B & DFN & - & 70.1 & 83.0 & - & 54.1 & 58.2 & - & 84.3 \\
        \quad SlowFast-LLaVA-1.5-7B & Oryx-ViT & 2.36M & 67.5 & 80.4 & 91.1 & 49.0 & 62.5 & 76.4 & 76.4 \\
        \arrayrulecolor{lightgray}\cmidrule(rl){1-10}
        \quad SlowFast-LLaVA-1.5-7B & \atoken-So/\textsc{c} & 2.36M & 68.8 & 81.2 & 92.1 & 48.7 & 61.2 & 74.5 & 77.7 \\
\arrayrulecolor{black}\bottomrule
\end{tabular}
}
    \vspace{-3mm}
\end{table*}

%% file: table/video_und.tex
\begin{table*}[!t]
\centering
\caption{
\small \textbf{Video understanding performance on multimodal LLMs.} 
Evaluation of SlowFast-LLaVA-1.5 with frozen \atoken-So/\textsc{c} vision encoder versus Oryx-ViT and other video MLLMs. Results shown for 6 benchmarks (general and long-form video understanding) across 1B, 3B, and 7B model scales.
}
\label{tab:video_understanding_results}
\resizebox{\textwidth}{!}{
\begin{tabular}{l cc cccc ccc}
\toprule
& \multirow[b]{3}{*}{\begin{tabular}{c}\textbf{Vision}\\\textbf{Encoder}\end{tabular}} & \multirow[b]{3}{*}{\begin{tabular}{c}\textbf{\# Input}\\\textbf{Tokens}\end{tabular}} & \multicolumn{3}{c}{\textbf{General VideoQA}} & \multicolumn{3}{c}{\textbf{Long-Form Video Understanding}} \\
\cmidrule(lr){4-6} \cmidrule(lr){7-9}
& & & \textbf{VideoMME} & \textbf{PercepTest} & \textbf{NExT-QA} & \textbf{LongVideoBench} & \textbf{MLVU} & \textbf{LVBench} \\
\textbf{Multimodal LLM} & & & \textbf{(w/o sub)} & \textbf{(val)} & \textbf{(test)} & \textbf{(val)} & \textbf{(m-avg)} & \textbf{(avg)} \\
\midrule
\rowcolor{gray!15} \multicolumn{9}{l}{\textit{\textbf{1B Model Comparison}}} \\
        \quad Apollo-1.5B & SigLIP & 3K & 53.0 & 61.0 & - & 54.1 & 63.3 & - \\
        \quad InternVL2.5-2B & InternViT & 16K & 51.9 & - & 77.2 & 52.0 & 61.4 & 37.9 \\
        \quad Qwen2-VL-2B & DFN & 16K & 55.6 & 53.9 & 77.2 & 48.7 & 62.7 & 39.4 \\
        \quad SlowFast-LLaVA-1.5-1B & Oryx-ViT & 9K & 56.6 & 61.9 & 76.7 & 54.3 & 64.3 & 39.7 \\
        \arrayrulecolor{lightgray}\cmidrule(rl){1-9}
        \quad SlowFast-LLaVA-1.5-1B &  \atoken-So/\textsc{c} & 9K & 56.7 & 63.9 & 74.8 & 55.1 & 64.7 & 41.1 \\

\arrayrulecolor{black}\midrule
\rowcolor{gray!15} \multicolumn{9}{l}{\textit{\textbf{3B Model Comparison}}} \\
        \quad InternVL2-4B & InternViT & 16K & 53.9 & 53.9 & 71.1 & 53.0 & 59.9 & 35.1 \\
        \quad LinVT-Blip3-4B & SigLIP & - & 58.3 & - & 80.1 & 56.6 & 67.9 & - \\
        \quad Apollo-3B & SigLIP & 3K & 58.4 & 65.0 & - & 55.1 & 68.7 & - \\
        \quad SF-LLaVA-1.5-3B & Oryx-ViT & 9K & 60.8 & 65.8 & 80.8 & 57.2 & 68.8 & 43.3 \\
        \arrayrulecolor{lightgray}\cmidrule(rl){1-9}
        \quad SF-LLaVA-1.5-3B &  \atoken-So/\textsc{c} & 9K & 60.4 & 66.0 & 80.8 & 57.2 & 66.7 & 41.3 \\

\arrayrulecolor{black}\midrule
\rowcolor{gray!15} \multicolumn{9}{l}{\textit{\textbf{7B Model Comparison}}} \\
        \quad Oryx1.5-7B & Oryx-ViT & 14K & 58.8 & 70.0 & 81.8 & 56.3 & 67.5 & 39.0 \\
        \quad LLaVA-Video-7B & SigLIP & 11K & 63.3 & 66.9 & 83.2 & 58.2 & 70.8 & - \\
        \quad Apollo-7B & SigLIP & 3K & 61.3 & 67.3 & - & 58.5 & 70.9 & - \\
        \quad InternVL2.5-8B & InternViT & 16K & 64.2 & - & 85.0 & 60.0 & 69.0 & 43.2 \\
        \quad Qwen2-VL-7B & DFN & 16K & 63.3 & 62.3 & 81.2 & 55.6 & 69.8 & 44.7 \\
        \quad SlowFast-LLaVA-1.5-7B & Oryx-ViT & 9K & 63.9 & 69.6 & 83.3 & 62.5 & 71.5 & 45.3 \\
        \arrayrulecolor{lightgray}\cmidrule(rl){1-9}
        \quad SlowFast-LLaVA-1.5-7B & \atoken-So/\textsc{c} & 9K & 64.5 & 70.3 & 83.7 & 60.6 & 69.8 & 44.8 \\
\arrayrulecolor{black}\bottomrule
\end{tabular}
}
    \vspace{-3mm}
\end{table*}

%% file: table/imagenet_gen.tex
\begin{table}[t]
    \centering
    \caption{\small
        \textbf{Class-conditional image generation on ImageNet 256x256.} We compare different \atoken stages against the specialized VAVAE tokenizer using the Lightning-DiT framework. We report gFID, sFID, Inception Score (IS), Precision (Pre.), and Recall (Rec.). 
        $^\dagger$The VAVAE baseline applies CFG only to the first 3 latent channels, while we follow the standard protocol of applying it to all channels.
    }
    \label{tab:main_image_t2i}
      \resizebox{0.75\columnwidth}{!}{
    \begin{tabular}{l c c c c c c c}
        \toprule
        \textbf{Tokenizer} & \begin{tabular}{c}\textbf{Latent}\\\textbf{Channels}\end{tabular}  & \begin{tabular}{c}\textbf{CFG}\\\textbf{Scale}\end{tabular} & \textbf{gFID}$\downarrow$ & \textbf{sFID}$\downarrow$ & \textbf{IS}$\uparrow$ & \textbf{Pre.}$\uparrow$ & \textbf{Rec.}$\uparrow$ \\
        \midrule
        DiT & 4 & 1.5 & 2.27 & 4.60 & 278.2 & \textbf{0.83} & 0.57\\
        SiT & 4 & 1.5 & 2.06 & 4.50 & 270.3 & 0.82 & 0.59 \\
        REPA & 4 & 1.35 & 1.42 & 4.70 & \textbf{305.7} & 0.80 & \textbf{0.65} \\
        VAVAE & 32 & 6.7$^\dagger$ & \textbf{1.35} & \textbf{4.15} & 295.3 & 0.79 & \textbf{0.65} \\
        \arrayrulecolor{black}\cmidrule(lr){1-8}
        % \multicolumn{7}{l}{\cellcolor{gray!15}\atoken ~\textit{base}} \\
        \multicolumn{7}{l}{\atoken-B/\textsc{c}} \\
        \quad Stage 1 & 32 & 1.5 & 1.44 & 4.71 & 273.3 & 0.79 & 0.64 \\
        \quad Stage 2 & 48 & 1.65 & 1.54 & 4.90 & 254.7 & 0.77 & \textbf{0.65} \\
        \quad Stage 3 & 48 & 1.65 & 1.58 & 4.86 & 254.6 & 0.76 & \textbf{0.65} \\
        \arrayrulecolor{black}\cmidrule(lr){1-8}
        \multicolumn{7}{l}{\atoken-So/\textsc{c}} \\
        \quad Stage 1 & 32 & 1.5 & 1.62 & 4.54 & 253.3 & 0.78 & 0.63 \\
        \quad Stage 2 & 48 & 1.65 & 1.88 & 4.71 & 231.1 & 0.80 & 0.60 \\
        \quad Stage 3 & 48 & 1.65 & 1.56 & 4.60 & 260.0 & 0.79 & 0.63 \\
        \arrayrulecolor{black}\bottomrule

    \end{tabular}
    }
\end{table}

%% file: table/imagenet_gen_discrete.tex
\begin{table}[h!]
    \centering
\caption{\small
    \textbf{Discrete Tokenizer Class-conditional Image Generation on ImageNet.} 
    We evaluate \atokensod~against other discrete tokenizer-based generation models. 
    Metrics include model parameters, CFG scale, gFID, Inception Score (IS), Precision, and Recall.
}
    \label{tab:main_image_t2i_discrete}
    \resizebox{0.75\columnwidth}{!}{
    \begin{tabular}{llcccccc}
        \toprule
        \textbf{Tokenizer} & \textbf{Generator} & \textbf{\# Params} & \begin{tabular}{c}\textbf{CFG}\\\textbf{Scale}\end{tabular} & \textbf{gFID}$\downarrow$ & \textbf{IS}$\uparrow$ & \textbf{Pre.}$\uparrow$ & \textbf{Rec.}$\uparrow$ \\
        \midrule
        LFQ & MAGVIT-V2 & 307M & - & 1.91 & \textbf{324.3} & - & - \\
        TikTok-L & MaskGiT & 227M & - & 6.18 & 182.1 & 0.80 & 0.51 \\
        VQGAN & LlamaGen & 1.4B & 1.75 & 2.34 & 253.9 & 0.81 & 0.60 \\
        UniTok & LlamaGen & 1.4B & 1 & 2.51  & 216.7 & \textbf{0.82} & 0.57 \\
        TokenBridge & TokenBridge-L & 486M & 3.1 & \textbf{1.76} & 294.8 & 0.80 & \textbf{0.63} \\
        \midrule
        \atokensod & TokenBridge-L & 548M & 3.1 & 2.23 & 274.5 & 0.79 & 0.61 \\
        \bottomrule
    \end{tabular}
    }
\vspace{-3mm}
\end{table}

%% file: table/video_gen.tex
\begin{table*}[!t]
\centering
\caption{\small \textbf{Text-to-image and text-to-video generation benchmarks.} 
We compare \atoken~Stages 2-3 with specialized video tokenizers (Cosmos, Hunyuan, Wan) under resource-constrained settings. Higher scores indicate better performance across all metrics.
All models trained with identical data and model sizes for fair comparison.}

\label{tab:video_generation_results}
\resizebox{\textwidth}{!}{
\begin{tabular}{l ccc ccc ccc}
\toprule
 & \multirow[b]{2}{*}{\begin{tabular}{c}\textbf{Comp.}\\\textbf{Ratio}\end{tabular}} & \multirow[b]{2}{*}{\begin{tabular}{c}\textbf{Latent}\\\textbf{Size}\end{tabular}} & 
 \multirow[b]{2}{*}{\begin{tabular}{c}\textbf{Patch}\\\textbf{Size}\end{tabular}} & \multicolumn{3}{c}{\textbf{T2I}} & \multicolumn{3}{c}{\textbf{T2V: VBench}} \\
\cmidrule(lr){5-7} \cmidrule(lr){8-10}
\textbf{Tokenizer} & & & & \textbf{CLIP} & \textbf{Pick} & \textbf{GenEval} & \textbf{Quality} & \textbf{Semantic} & \textbf{Total} \\
\midrule
Cosmos-0.1-CV4×8×8 & (4,~~8,~~8)   & 16 & 2 & 32.16 & 21.47 & 62.14\% & 77.27\% & 65.13\% & 74.84\% \\
Hunyuan & (4,~~8,~~8)   & 16 & 2& 32.49  & 21.66 & 66.11\% & 79.52\% & 72.03\% & 78.02\% \\
Wan2.1 & (4,~~8,~~8) & 16 & 2 & 32.45  & 21.62 & 65.57\% & 79.74\% & 74.01\% & 78.60\% \\
% Wan2.2      & (4,16,16) & 1 & ?  & -       & -       & -      & -     & -     & - \\
\midrule
% --- Styling our model to stand out, similar to the reference ---
\multicolumn{7}{l}{\atoken-So/\textsc{c}} \\
        \quad Stage 2 & (4,16,16) & 48 & 1 & 32.44 & 21.59 & 63.08\% & 79.30\% & 72.42\%  & 77.92\% \\
        \quad Stage 3 & (4,16,16) & 48 & 1 & 32.50 & 21.74 & 64.61\% & 79.82\% & 73.04\%  & 78.46\% \\
\bottomrule
\end{tabular}
}
\vspace{-3mm}
\end{table*}

%% file: sec/6_relate.tex
\section{Related Work}
\label{sec:related_work}
\paragraph{Reconstruction Tokenizers.}
% Image
High-resolution images have been compressed using deep auto-encoders \citep{hinton2012improving, vincent2008extracting}, which learn lower-dimensional latent representations for reconstruction.
VAEs \citep{kingma2013auto} extended this framework with probabilistic modeling, while VQ-VAE \citep{van2017neural} introduced vector quantization to discretize the latent space. Building on these foundations, subsequent works enhanced reconstruction quality through adversarial training \citep{rombach2022high, esser2020taming}, developed alternative quantization strategies \citep{Lee2022AutoregressiveIG, Mentzer2023FiniteSQ, Luo2024OpenMAGVIT2AO, Zheng2022MoVQMQ}, incorporated semantic guidance \citep{Li2024ImageFolderAI, Li2024XQGANAO, Yao2025ReconstructionVG, Zha2024LanguageGuidedIT, Chen2024SoftVQVAEE1, Chen2025MaskedAA, Kim2025DemocratizingTM}, and scaled model capacity \citep{xiong2025gigatok}.

% Video
Video tokenization extended these image-based methods to temporal domains, employing 3D convolutions \citep{yan2021videogpt, ge2022long, yu2023magvit}, decoupled spatial-temporal processing \citep{polyak2024movie}, and causal modeling \citep{kong2024hunyuanvideo, wan2025, Yang2024CogVideoXTD}.
% 3D: % Transformer
Beyond convolutional architectures, recent work has explored Vision Transformers \citep{dosovitskiy2020image} as an alternative backbone for both image \citep{yu2021vector, yu2024image, hansen2025learnings} and video \citep{villegas2022phenaki, wang2024omnitokenizer, wang2024larp, Yan2024ElasticTokAT} tokenization.

3D generation methods initially applied diffusion models directly to various 3D representations \citep{luo2021diffusion, hui2022neural, shue20233d, wang2023rodin, he2024gvgen}, then shifted toward compact latent spaces for improved efficiency \citep{gupta20233dgen, xiong2024octfusion, jun2023shap, lan2024ln3diff, nichol2022point}. Notably, Trellis \citep{xiang2024structured} introduces structured latents (SLAT) that jointly encode geometry and appearance on sparse 3D grids, enabling flexible decoding to multiple output formats.

\paragraph{Visual Encoders.}
% Image
Image encoders initially leveraged contrastive learning through vision-language alignment \citep{radford2021learning, jia2021scaling, Zhai2023SigmoidLF} and image-only self-supervision \citep{chen2020simple, oquab2023dinov2}.
Generative pretraining explored text generation objectives \citep{wang2021simvlm}, discrete token reconstruction \citep{bao2021beit}, and masked image modeling \citep{he2022masked, Carreira2024Scaling4R}. Methods like NaViT \citep{dehghani2023patch} introduced resolution flexibility with preserved aspect ratios. Recent unified approaches merge contrastive, generative, and self-supervised objectives \citep{yu2022coca, tschannen2025siglip} or leverage intermediate-layer features with task-specific alignment \citep{bolya2025perception}.
 
% Video
Video encoders primarily employ self-supervised learning on video-only data \citep{qian2021spatiotemporal, feichtenhofer2021large, Recasens2021BroadenYV, Qian2022OnTG, Tong2022VideoMAEMA} or video-language modeling with noisy text supervision \citep{Fu2021VIOLETE, Zellers2022MERLOTRN, Li2022LAVENDERUV, Huang2022CloverTA, Chen2023VideoLLMMV}.
Recent methods treat video as image sequences, focusing on context window expansion \citep{team2024gemini, Xue2024LongVILASL} or token compression \citep{Li2023LLaMAVIDAI, Song2023MovieChatFD, Fei2024VideoCCAMEV, Weng2024LongVLMEL, xu2024slowfast}.

% 3D
% \todo{3D encoder}

\paragraph{Unified Tokenizers \& Multimodal Models.}

Unified Multimodal Models aim to combine visual understanding and generation within a single framework \citep{Wang2022OFAUA, Mizrahi20234MMM, lu2024unified}.
Many approaches use decoupled tokenizers while employing various generation paradigms -- autoregressive \citep{Lu2022UnifiedIOAU, Team2024ChameleonME, Wu2024JanusDV}, diffusion \citep{Zhou2024TransfusionPT}, flow-matching \citep{Ma2024JanusFlowHA}, and masked prediction \citep{Xie2024ShowoOS, tian2025unigen}. 
Recent efforts on unified tokenizers that handle both tasks include VILA-U \citep{wu2024vila}, which combines pixel reconstruction with contrastive learning in a single vision tower; SeTok \citep{wu2024towards}, which groups visual features into semantic units; UniTok \citep{ma2025unitok}, which uses multi-codebook quantization for enhanced expressiveness; and UniToken \citep{jiao2025unitoken}, which produces hybrid discrete-continuous representations through dual encoders. 
Show-o2 \citep{Xie2025Showo2IN} extends these approaches by leveraging a 3D causal VAE space with dual-path spatial-temporal fusion, enabling scalability across both image and video modalities while combining autoregressive modeling with flow matching.

%% file: sec/7_conclusion.tex
\section{Discussion and Conclusion}

The effectiveness of \atoken~across diverse modalities and tasks suggests new opportunities: 
visual tokenization can achieve the same unification that transformed language modeling. Our single framework achieves both high-fidelity reconstruction and semantic understanding across images, videos, and 3D assets. This integration became possible through the combination of our sparse 4D representation, transformer-based architecture, adversarial-free training strategy, and progressive multimodal curriculum. Due to limited computational resources, we could only test \atoken on separate downstream tasks. Building the comprehensive omnimodel that would demonstrate \atoken's full potential remains as future work.
Looking forward, \atoken~opens paths for visual foundation models to follow language modeling's trajectory toward true generalization. We hope this work sheds light on the next-generation multimodal AI systems built upon unified visual tokenization.

%% file: sec/supp_0_contribution.tex
\section{Contributions}
\label{appendix:contribution}

Jiasen designed the main concept and project scope, developed the unified representation, main architecture, native resolution training, distill-based semantic loss, stage-wise round robin training strategy, discrete quantization, and KV-cache video decoding, GAN training recipe \etc. Curated the image and video dataset, trained the model, conducted in-training evaluation, and wrote the paper.
Liangchen oversaw engineering aspects for the project, developed the sparse transformer structure, adversarial-free training loss recipe, video reconstruction loss recipe, and 3D tokenizer pipeline and dataset. Evaluated image reconstruction and understanding (\cref{subsec:image}), 3D reconstruction and understanding (\cref{subsec:3d}), ran image generation with continuous tokens (\cref{subsec:image_gen_continous}) and text-to-3D synthesis (\cref{subsec:text_to_3d}), and contributed to writing the paper.
Mingze contributed to the video understanding dataset design and suggested video understanding encoding settings.
% , and helped launch the training experiment.
Ran all Multimodal LLM experiments and wrote the corresponding section (\cref{subsec:mllm}).
Byeongjoo contributed to the discussion of GAN settings and video reconstruction frame sampling strategy. Evaluated video reconstruction (\cref{subsec:video}) and ran text-to-video generation experiments and wrote the corresponding section (\cref{subsec:text_to_video}).
Yanjun evaluated video retrieval (\cref{subsec:video}), ran image generation experiments with discrete tokens, and wrote the corresponding section (\cref{subsec:image_gen_discrete}).
Chen contributed to discussions on image understanding and image generation with continuous tokens.
Afshin advised on research direction and helped manage compute resources.
Yinfei advised on research direction, provided feedback through regular discussions, and helped manage computing resources.